\documentclass[
]{ceurart}

\usepackage{listings}
\usepackage{amsfonts}
\usepackage{amssymb}
\usepackage{amsthm}
\usepackage{amsmath}
\usepackage{latexsym}
\usepackage{stmaryrd}
\usepackage{wasysym}
\usepackage{mathrsfs}
\usepackage{graphicx}
\usepackage{rotating}
\usepackage{pifont}
\usepackage{booktabs}
\usepackage{array}
\usepackage{multirow}
\usepackage{tikz}
\usepackage{color}
\usepackage{xcolor}
\usepackage{enumitem}
\usepackage{xspace}
\usepackage{thm-restate} 
\usepackage{bussproofs}
\usepackage{tcolorbox}
\usepackage{svg}

\usepackage{subcaption}
\usepackage[capitalize]{cleveref}
\Crefname{subsection}{Subsection}{Subsections}
\crefname{subsection}{Subsection}{Subsections}

\usepackage{longtable}

\newtheorem{example}{Example}

\newcommand{\q}{\psi}

\newcommand{\emb}[1]{\mathbf{#1}}

\usepackage[T1]{fontenc}
\usepackage[utf8]{inputenc}

\begin{document}

\copyrightyear{2022}
\copyrightclause{Copyright for this paper by its authors.
  Use permitted under Creative Commons License Attribution 4.0
  International (CC BY 4.0).}

\conference{KR4HI'22: International Workshop on Knowledge Representation for Hybrid intelligence,
  June 14, 2022, Amsterdam}

\title{On the Effectiveness of Knowledge Graph Embeddings: a Rule Mining Approach}

%


\author[1]{Johanna J{\o}sang}[%
email=lisa.josang@student.uib.no,
]
\address[1]{Department of Informatics, University of Bergen}

\author[1]{Ricardo Guimarães}[%
orcid=0000-0002-9622-4142,
email=ricardo.guimaraes@uib.no,
url=https://rfguimaraes.github.io,
]

\author[1]{Ana Ozaki}[%
orcid=0000-0002-3889-6207,
email=ana.ozaki@uib.no,
url=https://www.uib.no/en/persons/Ana.Ozaki,
]



\begin{abstract}
We study the effectiveness of Knowledge Graph Embeddings (KGE)
for knowledge graph (KG) completion with rule mining. 
More specifically, we mine rules from KGs before and
after they have been completed by a KGE to compare 
possible differences in the rules extracted. We apply this method
to classical KGEs approaches, in particular, TransE, DistMult and ComplEx.
Our experiments indicate that there can be huge differences
between the extracted rules, depending on the KGE approach for KG completion. 
In particular, after the TransE completion,  several
spurious rules were extracted.   
\end{abstract}


\maketitle
%


\section{Introduction}

Nowadays, Knowledge Graphs (KGs) are an increasingly popular way to represent data~\cite{Hogan2021}.
A KG can be often seen as a labelled directed graph  in which the nodes represent the elements in the domain of interest (e.g. people) and the edges represent a relation between two elements.
For instance, a KG such as Wikidata, might include the node ``Paris'' with an outgoing edge labelled ``capital of'' to the node ``France''.

Knowledge graph embeddings (KGEs) attempt to capture patterns present 
in KGs and generalize them so as to infer new
data (commonly, in the format of RDF triples). Such data can then 
be used to ``complete'' the information in the KG. 
Since the publication of the very classical  KGE approach known as TransE~\cite{Bordes2013}, several
authors proposed alternative approaches for KGEs.
KGEs can be effective methods for KG completion if they  capture patterns
in the data and  can generalize such patterns in a uniform way. That is, 
there should not be much distortions on the facts classified as plausible 
by a KGE method.

A main challenge is to determine and evaluate how effective
they are for KG completion. 
In other words, how well such
KGEs can generalize from the data and 
whether there are biases in the process that
significantly affect KG completion based on such models. 
Understanding the power of KGEs is vital to enable human comprehension of the power and limitation of these models.
\Cref{ex:family} illustrates how a KGE can be used for KG completion.

\begin{example}\label{ex:family} If most individuals in the data who are siblings are also relatives, then we would expect a KGE approach to capture such patterns. The KGE would give a high rank to triples expressing two individuals known to be siblings as relatives.
\end{example}

However, the underlying \emph{rules} captured by   KGEs are 
hidden in the model. To study the effectiveness of KGEs for KG completion,
 we apply \emph{rule mining}: a data mining based approach for extracting rules that represent the main 
patterns in the data~\cite{Lajus2020}. 
Such rules can be useful to identify implicit knowledge in the 
domain associated with the data. 
The following \lcnamecref{ex:family_rule} demonstrates this.

\begin{example}\label{ex:family_rule} In the context of \cref{ex:family},
  such rule could be 
`if two persons are siblings then
they are (family) relatives''.
In symbols, $sibling(x,y)\Rightarrow relative(x,y)$ (as usual, e.g.~\cite{Lajus2020}, 
the universal quantifier over the variables has been omitted). 
\end{example}

\begin{figure}
\includegraphics[width=1\linewidth]{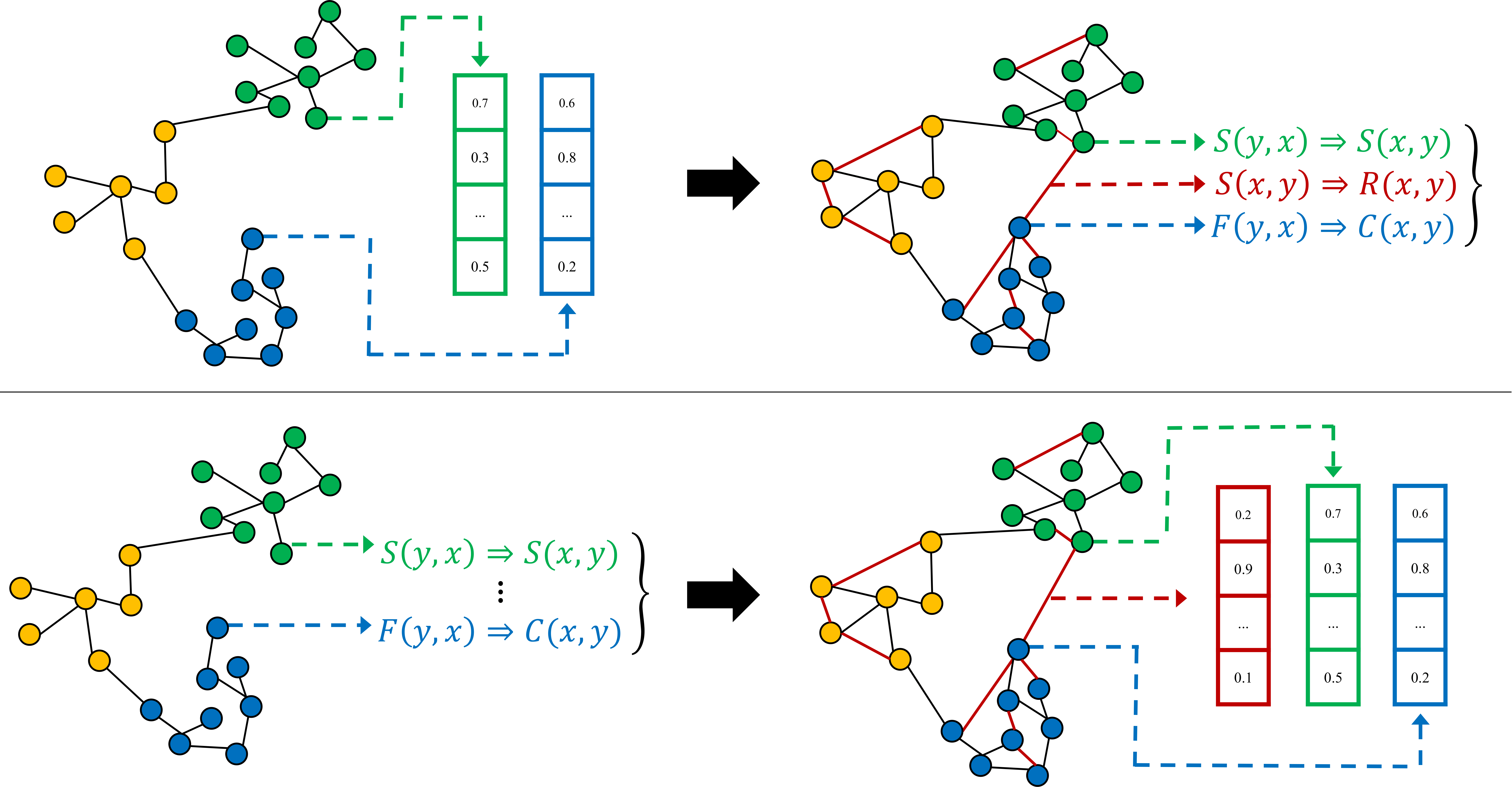}
    \caption{KGEs can be applied for KG completion and impact the rules that hold with high confidence in the extended (completed) KG. 
    Rule Mining can be applied to induce new links and change KGEs results, however, we focused on the first case since our goal is to investigate the effectiveness of KGEs.}%
\label{fig:kgerule}
\end{figure}

Rule mining and KGEs can complement each other
as methods for identifying patterns in KGs. 
\cref{fig:kgerule} illustrates 
the interplay between KGEs and rule mining.
In this work, we mine rules from KGs before and
after they have been completed by a KGE to compare 
possible differences in the rules extracted.  
As we discuss in this work, 
there can be significant distortions for KG completion depending on the chosen
KGE approach. 
Our work gives some insights about how to detect such distortions.
We apply our method
to classical KGEs approaches, in particular, TransE, DistMult and ComplEx.

\smallskip

In the next sections, we discuss related work (\cref{sec:related_work}), introduce the most relevant basic notions from the literature (\cref{sec:preli}), describe our method (\cref{sec:rule_mining_extended}),
present the experimental results (\cref{sec:experiments}), and conclude (\cref{sec:conclusion}).

\section{Related Work\label{sec:related_work}}

In this \lcnamecref{sec:related_work}, we discuss the approaches which relate more closely to this work.
We focus in particular on rule mining methods which also employ KGEs, or similar approaches.
KALE~\cite{Guo2016}, for example, learns rules in a two-step process: first by mining rules from TransE embeddings and then retraining the embeddings on a joint model for triples and rules.
While KALE unifies KGEs and rule learning, our approach is agnostic regarding the KGE model.
RUGE~\cite{Guo2018} is based on KALE and employs a similar procedure where soft rules are obtained from the data, which will guide the embeddings learned.
In both KALE and RUGE, only the embeddings are updated, the rules remain the same from the moment they are extracted.
In contrast, IterE~\cite{Wang2019} iteratively updates both embeddings and the rules.
The score of the rules in IterE depends directly on the embeddings of the relations that it uses. ExCut~\cite{10.1007/978-3-030-62419-4_13} also employs KGEs and rule mining, however, with the purpose of explaining clusterings of entities.

\Citet{Meilicke2018} combines rule-based approaches and KGEs using an \emph{ensemble approach}.
In ensemble learning, the idea is that each model provides an answer to a problem, and then these individual answers are aggregated to give a final verdict~\cite{Wang2018}.
Since rule-based and KGE-based methods often offer complimentary performance on different instances, the authors devised a system that performs well in different settings for KG completion~\cite{Meilicke2021}. 
The key distinction between the mentioned studies and ours is that they aim at finding true triples to perform KG completion, while we focus our investigation on  the effectiveness of KG completion via KGEs by mining rules before and after the completion and then analysing the results.

\section{Background}\label{sec:preli}

Here we provide basic notions about KGEs and rule mining relevant for this paper.

\subsection{KG Embeddings}

KGEs are functions that map entities and relations in KG to a vector space, usually with  low-dimension.
The main goal of KGE methods is to represent entities and relations in such a way that the patterns of the data are preserved.
In particular, one of the main applications of these methods is KG completion: the task of identifying missing triples of a KG.\@

The mathematical representation assigned to an entity and/or a relation is called its embedding and each KGE approach  determines  constraints on how the embeddings of entities and relations relate to each other.
For instance, in the TransE model~\cite{Bordes2013}, every entity and relation is mapped to a vector in \(\mathbb{R}^n\). 
Given a triple \((s, r ,o)\) in the KG, the TransE model aims at minimising the distance between \(\emb{s} + \emb{r}\) and \(\emb{o}\), where \(\emb{s}\), \(\emb{r}\) and \(\emb{o}\) are the embeddings of \(s\), \(r\) and \(o\)\footnote{The letters 
`s', `r', and 
`o' stand for \emph{subject}, \emph{relation}, and \emph{object}, respectively 
The letters
`h' and `t' are also often used instead of `s' and `o', meaning \emph{head} and \emph{tail}, respectively.}.

The computation of KGEs follows the standard 
machine learning pipeline: the embeddings start with randomly assigned values, which are then updated using stochastic gradient descent or similar measures while minimising a loss function that encodes the desired property imposed by the model.
The loss function usually depends on a score (or energy) function, which determines, for a given triple, how likely it is to be true, given the embeddings of its elements.
In the case of TransE, the score function is \(f_{\text{TransE}}(s, r, o) = {||\emb{s} + \emb{r} - \emb{o}||}_\ell\), in which \({||.||}_\ell\) denotes the \(\ell\)-norm (usually \(L_1\) or \(L_2\)-norm).

Nowadays, there are numerous approaches, including more sophisticated versions of TransE~\cite{Dai2020} which are also based on geometric formulations.
There are also approaches such as DistMult~\cite{Yang2015} which is based on tensor factorisation.
Embedding models vary concerning the number of parameters learned per entity and relation, and which patterns they can capture from data.
For instance, while DistMult is computationally cheap, it is unable to represent asymmetric relations such as \emph{sibling}, while ComplEx demands more parameters, but can represent both symmetric and asymmetric relations~\cite{Trouillon2016}.

\subsection{Rule Mining}\label{sec:rule-mining}

An \emph{atom} is an expression of the form $r(x, y )$, where $r$ is
a relation and $x, y$ are   variables.  
Rules in this work are expressions of the form 
$\mathbf{B}\Rightarrow H$ 
where $\mathbf{B}$ (called the \emph{body} of the rule) 
is a conjunction of atoms and $H$ is an atom (called the \emph{head} of the rule). A rule is \emph{closed} if all variables appear in at least two atoms. A closed
rule is always \emph{safe}, i.e. all head variables appear also in at least one body atom.
From now on, assume all rules we speak of are closed.
Regarding the measure used for ranking the rules, we use 
the \emph{Partial Completeness Assumption (PCA) confidence} (see Appendix~\ref{ap:pca} for details). This measure is more elaborate than 
the classical notion of confidence from rule mining in KGs~\cite{Lajus2020}, which 
is defined as the proportion of true predictions out
of the true   and false predictions. The PCA lies between
the open and the closed world assumptions. The main intuition is that if a node
is a parent of another node via a relation $r$ then it is assumed that all the information regarding childhood of 
this node by $r$ is complete. As an example, if we have the information that 
the mathematician Artur Ávila was born in Brazil, we assume that 
an assertion that he was born in Argentina is false.
On the other hand,  if we do not know where the mathematician Graciela Boente was born,
then we do not assume that an assertion that she was born  in Argentina is false.

\section{Comparing KGEs with Rule Mining\label{sec:rule_mining_extended}}

Here we describe our approach for investigating 
the effectiveness of KGEs for KG completion.
By KG completion we mean the process of adding new, potentially true triples to the original KG.\@
As the resulting KG may still be incomplete, we will refer to the new versions generated through this process as \emph{extensions} or \emph{extended KGs}.
In other words, we study how KGEs can be applied to
extend KGs with new triples which are associated with high PCA confidence.

We compare the results of rule mining 
before and after the KG has been extended by 
new triples. 
First we create the triples  
using an entity selection method and then 
check what is the confidence (w.r.t.\ the KGE). 
In summary, the main steps of our pipeline are as follows.
\begin{enumerate}
\item \textbf{Rule Mining on Original KG:} We apply rule mining to the KG.
\item\textbf{KG Completion:} We extend the  KG with different 
KGEs and entity selection methods.
\item\textbf{Rule Mining on Extended KG:} We apply rule mining to the extended KG.
\item\textbf{Analysis:} To study the effectiveness of the KG completion, we 
compare the rules mined before and after extending the KG.\@
\end{enumerate}

%

In Step 1, we apply PCA confidence (see definition in Appendix~\ref{ap:pca}) from rule mining.
In Step~2, we consider
three KGEs and three entity selection methods, which we detail in the next paragraph. 
Finally, in Steps 3 and 4 we apply rule mining in the extended KG
(using PCA confidence to score the rules)
and compare the rules. 
%


In Step 2 of our pipeline, we attempt to extend the KG 
with new triples that received a high ranking according 
to a KGE.\@ The challenge here is that it is not feasible to
check for all possible triples. In practice, one needs
to apply some heuristics to find ``good triples'', that is,
new triples that are good candidates to receive a high 
 ranking. 
To address this issue, we considered the strategies already implemented in 
AmpliGraph~\cite{luca_costabello_2021_4792436}.
However,
we had technical difficulties when trying to use them\footnote{
In the current version of AmpliGraph (1.4.0) it is not recommended to 
use  exhaustive search (e.g.\ for a random selection) 
 due to a large amount of computation 
required to evaluate all triples
\url{https://docs.ampligraph.org/en/1.4.0/generated/ampligraph.discovery.discover_facts.html}.
There were technical difficulties 
in using the other strategies.}.
The most effective and efficient strategy 
 employed by AmpliGraph to  
find ``good triples'' is to search for less frequent 
entities\footnote{The AmpliGraph team say that 
this assumption has been
true for their empirical evaluations, but is not necessarily true for all datasets~\cite{luca_costabello_2020_4268208}.}.
We 
implemented 
this strategy in this work.
To study the effect of frequency, we also 
implemented an entity selection method
that selects the most frequent and 
(uniform) random selection.
In addition, 
we included 
a probabilistic selection method based on the frequency of the entities in the dataset, where the
least frequent entities are most likely to be selected.
The latter served as a method between the least frequent and 
the random selection methods.



\section{Experiments\label{sec:experiments}}

In this \lcnamecref{sec:experiments}, we discuss the implementation of the pipeline and methodology presented in \cref{sec:rule_mining_extended} and present the results of the experiments. 
The main parameters of our experiments are 
the KGE, the entity selection method, and
the cutoff of the ranking of the rules
(a high ranking
is associated with low number, where $1$ is the highest rank). 
The model selection, candidate generation and extension were implemented in Python, relying mostly on AmpliGraph.
We mined the rules from the extended KGs using AMIE3~\cite{Lajus2020}.
More precisely, we used the following command to mine rules: \texttt{java -jar amie-milestone-intKB.jar -bias lazy -full -noHeuristics -ostd [TSV file]}, in which the TSV file contains the input KG.\@
The experiments were run on a server with 64 GB of RAM and an Intel Core i9-7900X\ 3.3GHz processor.
In the following, we discuss the other important aspects and results.

\subsection{Experimental Setup\label{sec:setup}}

Next, we discuss the datasets and KGE models that we employed in our experiments.

\subsubsection{Datasets and Model Selection\label{sec:data_model_selection}}

We considered two KGs as datasets in this paper. 
However, instead of using them directly, we restricted our attention to six types of relations in each KG, removing the triples that use other relations.
We imposed this constraint to cope with the volume of data, and to control the number of resulting rules.
Moreover, as the we have a fixed limit on the number of candidate triples, limiting the number of relations increases the overall rate of triples per relation, which  facilitates rule mining and gives a clearer picture of the influence of each embedding method.

The first dataset is the well-known WN18RR, a KG derived from WordNet and that contains roughly 93000 triples and 11 relations~\cite{Dettmers2018}. 
We choose this KG mostly due to its popularity as a benchmark dataset for KG completion and its relatively small number of relation types.

In this KG, the entities are sets of words called \emph{synsets} and the relations 
connect synsets depending on their meaning. 
We selected the six most frequent relations from this KG, more specifically: \(hypernym\), \(derivationally\ related\ form\), \(member\ meronym\), \(has\ part\), \(synset\ domain\ topic\ of\) and \(instance\ hypernym\).
As a result, our version of the WN18RR KG has 88227 triples, corresponding to 95\% of the original KG.

The second is the Family KG, a portion of Wikidata5M obtained from the PyKEEN library\footnote{\url{https://pykeen.readthedocs.io/en/stable/api/pykeen.datasets.Wikidata5M.html\#pykeen.datasets.Wikidata5M}}.
Wikidata5M is a KG based on Wikidata with over 20 million triples over 800 types of relations~\cite{Wang2019}.
As with WN18RR, Wikidata5M is also commonly employed when evaluating methods for KG completion.
We focused on relations of the ``family'' domain, that is: P22 (\(father\)), P25 (\(mother\)), P26 (\(spouse\)), P40 (\(child\)), P1038 (\(relative\)) and P3373 (\(sibling\)). 
From now on, we will use the natural language readings (e.g. \(child\)) instead of the original property codes (e.g.\ P40) for readability.
We selected this particular subset of properties because the structure of family relations is well-known and easy to understand.
Therefore, it will facilitate the analysis of the rules obtained according to the KGEs. 
After restricting  to these six relations, the Family KG contains around 250000 triples.


Due to limited computational resources, we used a random search (instead of grid search) over all combinations of values for the hyperparameters.
Additionally, for each hyperparameter combination, we selected randomly and uniformly a learning rate between $10^{-4}$ and $10^{-2}$.
\Cref{tab:model_selection} depicts how the resulting models of each KG embedding method performed on each dataset regarding KG completion. 
We evaluated the different embedding methods on standard metrics for KG completion: mean rank (MR), the lower the better; mean reciprocal rank, which ranges from 0 (worst) to 1 (best); and hits@k which goes from 0 (worst) to 1 (best).


\begin{table}[thbp]
\centering
\begin{tabular}{@{}lrrrrrrrrrr@{}}
\toprule
\multicolumn{1}{c}{Dataset} & \multicolumn{5}{c}{WN18RR KG}                                           & \multicolumn{5}{c}{Family KG}                                           \\ \midrule
\multirow{2}{*}{Model}      & \multirow{2}{*}{MR} & \multirow{2}{*}{MRR} & \multicolumn{3}{c}{Hits@K}       & \multirow{2}{*}{MR} & \multirow{2}{*}{MRR} & \multicolumn{3}{c}{Hits@K} \\ \cmidrule(lr){4-6} \cmidrule(l){9-11} 
                            &                     &                      & 1            & 3                 & 10              &               &                      & 1             & 3       & 10     \\ \midrule
Random                      & 495.32              & 0.01                 & 0.00         & 0.00              & 0.01            & 498.72        & 0.00                 & 0.00          & 0.00    & 0.10   \\
TransE                      & \textbf{34.29}      & 0.60                 & 0.51         & \textbf{0.66}     & \textbf{0.76}   & \textbf{2.59} & 0.93                 & 0.88          & 0.97    & \textbf{0.99}   \\
DistMult                    & 152.37              & \textbf{0.62}        & \textbf{0.59}& 0.63              & 0.66            & 7.45          & 0.98                 & \textbf{0.99} & \textbf{0.99}    & \textbf{0.99}   \\
ComplEx                     & 139.36              & 0.59                 & 0.57         & 0.60              & 0.63            & 4.64          & \textbf{0.99}        & 0.98          & \textbf{0.99}    & \textbf{0.99}   \\ \bottomrule
\end{tabular}
\caption{Results of selected models evaluated on test set. The best results per column are highlighted.}%
\label{tab:model_selection}
\end{table}

In \cref{tab:model_selection}, we can see that the Random Baseline performed poorly in all metrics for every dataset.
This method simply assigns to each triple a pseudo-random number between 0 and 1, chosen uniformly.
While TransE has a number of limitations with relation that are symmetric or one-to-many, it was the best method regarding mean rank and Hits@10 on both datasets.
DistMult and ComplEx performed similarly in both datasets, even though both include asymmetric relations, which DistMult struggles to represent.

\subsubsection{KG Completion}

We generated one extension of each dataset for each of the 48 combination of parameters: 4 embedding models \(\times\) 4 candidate selection strategies \(\times\) 3 rank cutoff values.
Also, we created each extension in two steps: candidate triple generation and candidate ranking.

In the first step, we generate the triples that are going to be added to the original dataset.
Given one of the four entity selection methods we discussed in \cref{sec:rule_mining_extended}, we select the top 1000 entities according to the method.
For example, if the extension has \emph{least frequent} as its candidate selection method, we picked the 1000 least frequent entities in the original KG.\@
We limited the number of entities to 1000 due to computational limitations.
As a consequence of this limit, we need to consider at least $10^3 \times 6 \times 10^3 = 6 \times 10^6$ distinct triples for each extension, among those already in the original KG and the new candidates.

In the second step, we use the KGE models to decide which of the candidate triples should be added in the extension of a KG.\@
As the score functions in KG embedding are more useful for comparison rather than for deciding the plausibility of single triple in general, it is difficult to set an acceptance threshold based on score alone.
Therefore, we decided to use the rank of a triple instead.
Precisely, each extension has an associated value for the rank cutoff: if a candidate triple had a worse (higher) rank than the cutoff, it was not included in the extension.
The values which we considered, from the most to the least strict were 1, 4 and 7 because they linger in the usual ranges adopted for the Hits@K metric, which is commonly applied to evaluate performance in KG completion.

\subsection{\label{sec:results}Results}

In this \lcnamecref{sec:results}, we discuss the results of the experiments.
We focus on the contrast in number and quality of rules mined according to the embedding model, as it impact the results much more significantly, and later we analyse the effect of the other parameters. 


\begin{table}[thbp]
\centering
\begin{tabular}{@{}lrrrrrr@{}}
\toprule
Dataset         & \multicolumn{3}{c}{WN18RR}                                                                                                                                                                                                                                 & \multicolumn{3}{c}{Family}                                                                                                                                                                                                                                 \\ \cmidrule(l){2-7} 
Model           & \multicolumn{1}{c}{\begin{tabular}[c]{@{}c@{}}Orig.\ Rules \\ Found\end{tabular}} & \multicolumn{1}{c}{\begin{tabular}[c]{@{}c@{}}Orig.\ Rules \\ Missed\end{tabular}} & \multicolumn{1}{c}{\begin{tabular}[c]{@{}c@{}}New Rules\\ Found\end{tabular}} & \multicolumn{1}{c}{\begin{tabular}[c]{@{}c@{}}Orig.\ Rules \\ Found\end{tabular}} & \multicolumn{1}{c}{\begin{tabular}[c]{@{}c@{}}Orig.\ Rules \\ Missed\end{tabular}} & \multicolumn{1}{c}{\begin{tabular}[c]{@{}c@{}}New Rules\\ Found\end{tabular}} \\ \midrule
Random & 9                                                                                   & 1                                                                                    & 0                                                                             & 85                                                                                  & 9                                                                                    & 0                                                                             \\
TransE          & 10                                                                                  & 0                                                                                    & 659                                                                           & 94                                                                                  & 0                                                                                    & 986                                                                           \\
DistMult        & 10                                                                                  & 0                                                                                    & 5                                                                             & 93                                                                                  & 1                                                                                    & 90                                                                            \\
ComplEx         & 10                                                                                  & 0                                                                                    & 7                                                                             & 94                                                                                  & 0                                                                                    & 43                                                                            \\ \bottomrule
\end{tabular}
\caption{Distribution of rules over the extended KGs. Rules mined from multiple extensions count only once}%
\label{tab:nrules}
\end{table}

\Cref{tab:nrules} depicts the number of unique rules mined per embedding model considering all combinations of parameters on the different datasets.
There were no new rules obtained from KGs extended via the random baseline, which is expected, as it effectively adds triples randomly. 
Additionally, the random baseline was the worst regarding the extraction of original rules: one of the ten original rules of the WN18RR KG and 9 out of 94 for the Family KG were never mined from extensions with random baseline.
TransE extensions resulted in a large number of rules, when compared with the other methods, the number of new rules is over an order of magnitude higher than DistMult or ComplEx.
Also, when using DistMult-extended KGs, AMIE3 could not find the rule \(child(x, y) \Rightarrow mother(x, y)\) in the Family KG.\@
Even so, DistMult enabled more new rules than ComplEx on both KGs.
This difference is a consequence of ComplEx adding fewer candidate entities when compared with DistMult.
For instance, when using the \emph{probabilistic} selection method, DistMult assigned rank 1 to 1468 candidates on the WN18RR KG, while ComplEx gave rank 1 to only 866.
These results show that extending KGs in this way can give access to new rules, while preserving the old ones.
Furthermore, \cref{tab:nrules} conforms with our intuition that the differences between the results with DistMult and ComplEx should be smaller in comparison to the difference to the results with TransE.

\begin{figure}[htbp]
\centering
\begin{subfigure}{.5\textwidth}
  \centering
  \includegraphics[width=1\linewidth]{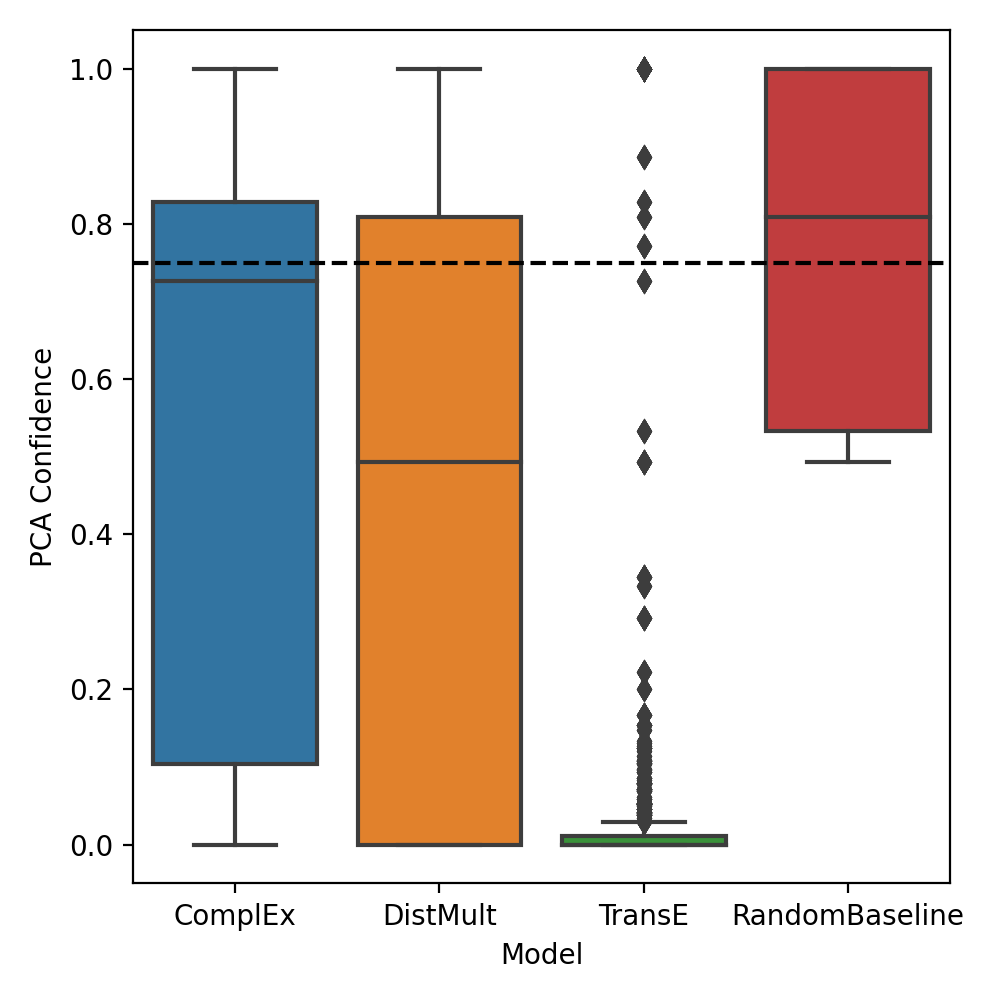}
  \caption{Original WN18RR KG}%
  \label{fig:PCA-models_wn18rr_boxplot_sub}
\end{subfigure}%
\begin{subfigure}{.5\textwidth}
  \centering
  \includegraphics[width=1\linewidth]{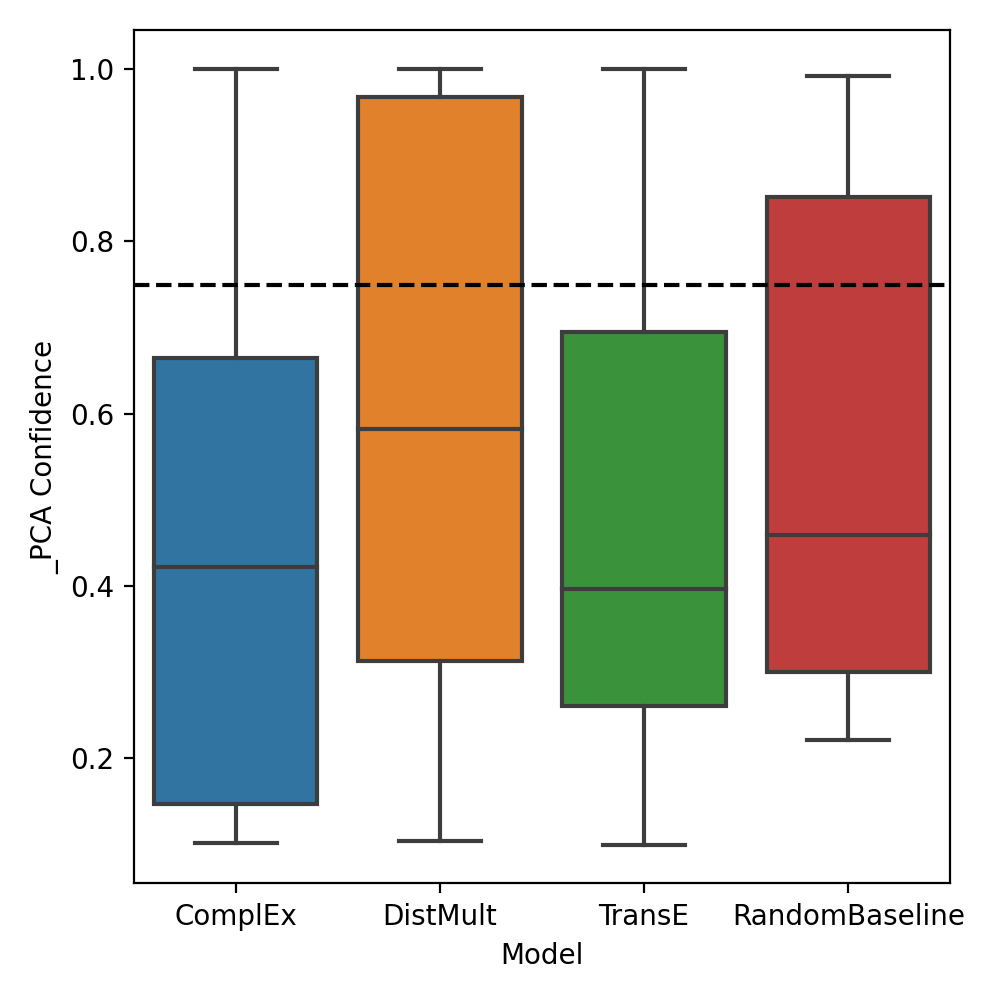}
  \caption{Extended WN18RR KG}%
  \label{fig:_PCA_models_wn18rr_boxplot_sub}
\end{subfigure}
\begin{subfigure}{.5\textwidth}
  \centering
  \includegraphics[width=1\linewidth]{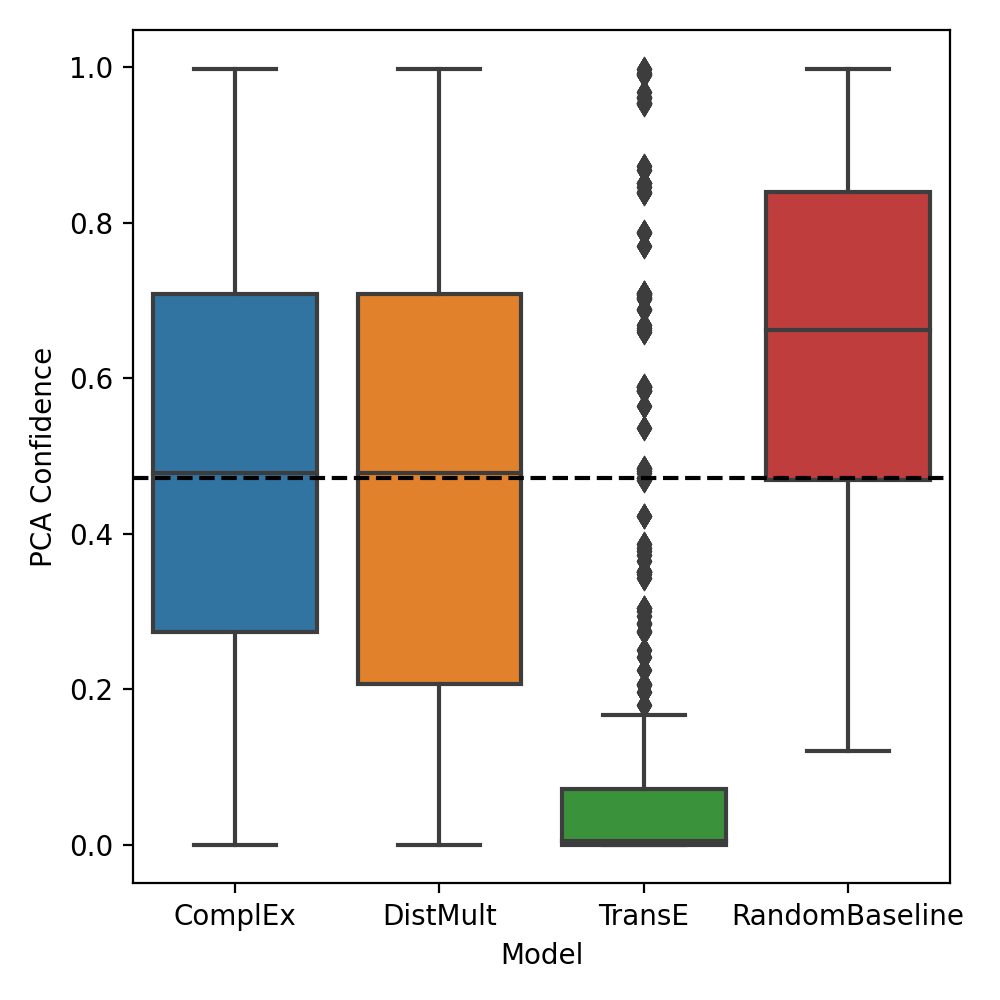}
  \caption{Original family KG}%
  \label{fig:models_family_boxplot_sub}
\end{subfigure}%
\begin{subfigure}{.5\textwidth}
  \centering
  \includegraphics[width=1\linewidth]{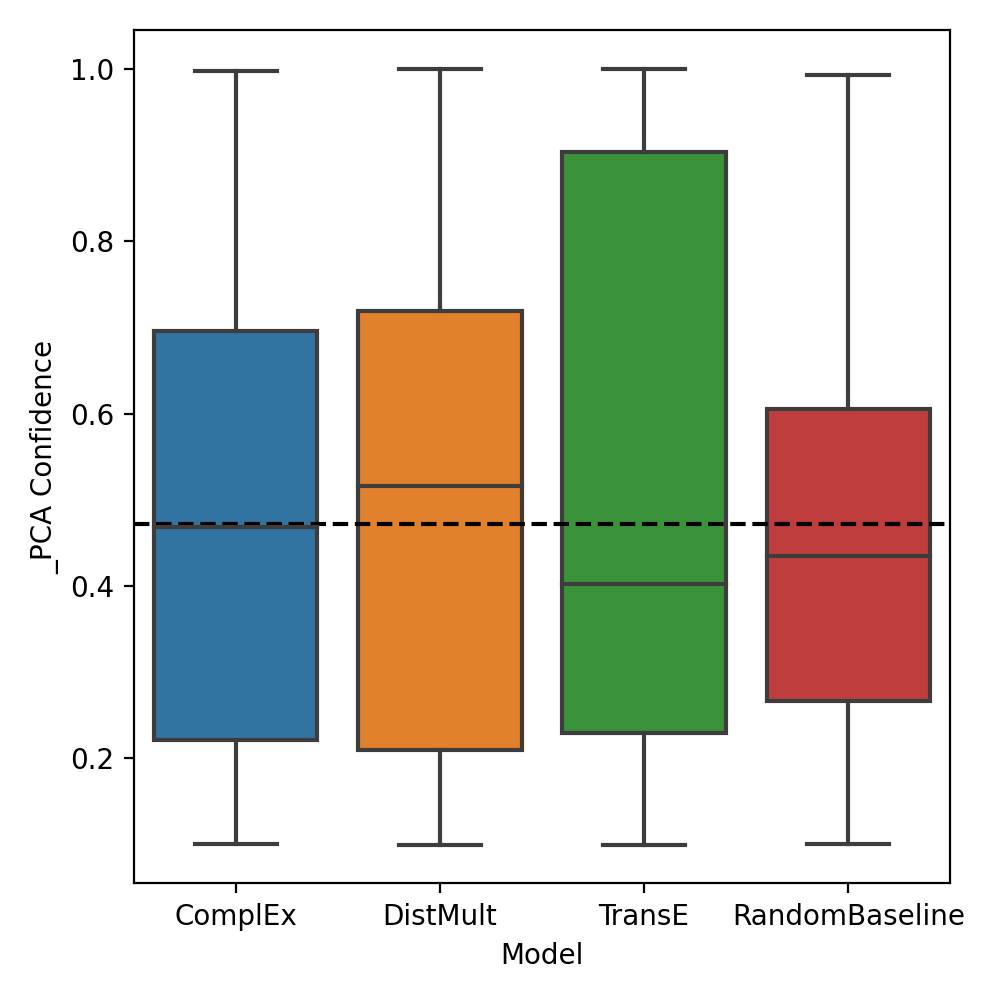}
  \caption{Extended family KG}%
  \label{fig:_PCA_models_family_boxplot_sub}
\end{subfigure}
\caption[Dist.\ of PCA conf.\ rules by KG embedding models]{\label{fig:PCA_models_boxplot}Distribution of PCA confidence of mined rules by KG embedding models. PCA confidence scores are calculated on the original KG and the extended KGs from which the rules are mined. The dashed line represents the median PCA confidence of the rules mined from the original KG.}
\end{figure}

Next, we discuss the PCA confidence of the rules mined.
\Cref{fig:PCA_models_boxplot} displays the distribution of PCA confidence of the rules discovered in the extensions, grouped by embedding method. The plots contain standard boxplots with whiskers indicating the variability outside the upper and lower quartiles, as well as outliers. The dashed horizontal line denotes the median of the PCA confidence for the original rules.
\Cref{fig:PCA-models_wn18rr_boxplot_sub,fig:models_family_boxplot_sub} consider the PCA confidence calculated on the original KGs, therefore those boxplots do not need to consider duplicates. 
In \cref{fig:_PCA_models_wn18rr_boxplot_sub,fig:_PCA_models_family_boxplot_sub} instead, each occurrence of a rule is treated as a separate instance (therefore, the same rule may be counted multiple times, if it was mined in different extensions of the same method).
The random baseline method produced a better PCA distribution than the others, but without new rules.
In contrast, with TransE produced a large number of rules, however most of these rules have low PCA values, as shown in \cref{fig:PCA_models_boxplot}.


In the Family KG, DistMult and ComplEx has similar performance regarding PCA confidence, with a slight lead by DistMult.
On the WN18RR the difference is more pronounced: on the original data, ComplEx has better PCA confidence (the median is much higher than for DistMult in \cref{fig:PCA-models_wn18rr_boxplot_sub}), in contrast, DistMult has better performance on its respective extensions (as shown by the overall position of the boxplots in \cref{fig:_PCA_models_wn18rr_boxplot_sub}).
As the PCA confidence in the extensions are biased towards the method themselves, this may indicate that ComplEx had a better performance than DistMult.

To understand what happens in the TransE case, let us consider the family dataset.
Since the relations \emph{sibling}, \emph{spouse}, and \emph{relative} are symmetric, they all collapse to the null vector.
Furthermore, in this dataset there is nothing that differentiates the relation \emph{mother} from \emph{father}, thus they are mapped to the same vector. 
As \emph{father} and \emph{mother} have the same embedding and are both inverses of \emph{child}, the embedding for \emph{child} is the opposite vector of the former.
Therefore, while TransE ranked existing triples well, as shown in the results for MR and MRR, it probably ranked many wrong triples as correct, generating a poor and biased set of candidate entities.
As a result, AMIE3 mined the rules \(spouse(x, y)  \Rightarrow father(x, y)\) and \(child(z, x) \wedge mother(z,y) \Rightarrow spouse(x, y)\) solely from TransE-extended KGs.
The resulting embeddings, computed in practice (see \cref{TransE_embedding_family,TransE_embedding_WN18RR} in the appendix) reflect this situation.
Not only that, although TransE mined many more rules, the mean PCA confidence of rules mined from TransE extensions is around 0.1, while the mean is much closer to 0.5 with the other embedding models.
That is not to say that all rules mined with TransE were bad or that all other methods produced plausible rules exclusively.
For instance, both TransE and DistMult extensions enabled the rule \(mother(x, y) \Rightarrow child(x,y)\).
In the DistMult case, this is likely a consequence of the fact that \(mother(x, y)\) and \(mother(y, x)\) will always be equality plausible according to DistMult (and the same happens for \(child(x, y)\) and \(child(y, x)\)).

\begin{figure}[thbp]
    \centering
\begin{subfigure}{.5\textwidth}
    \centering
    \includegraphics[width=0.9\linewidth]{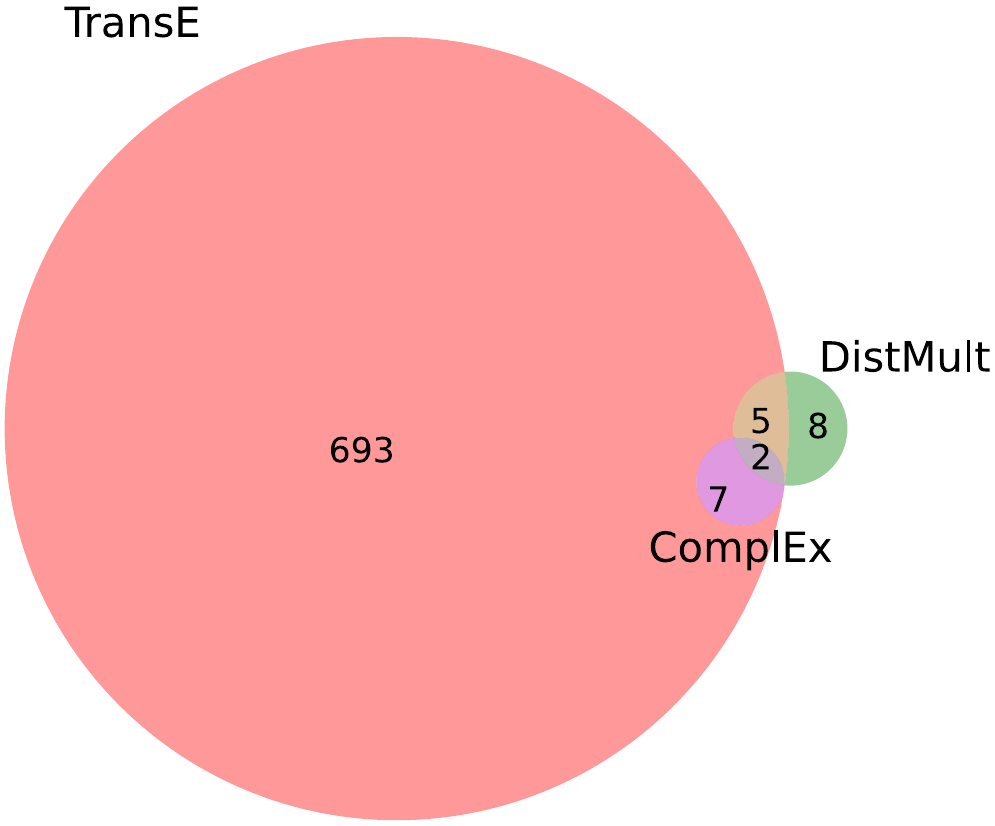}
\end{subfigure}%
\begin{subfigure}{.5\textwidth}
    \centering
    \includegraphics[width=0.9\linewidth]{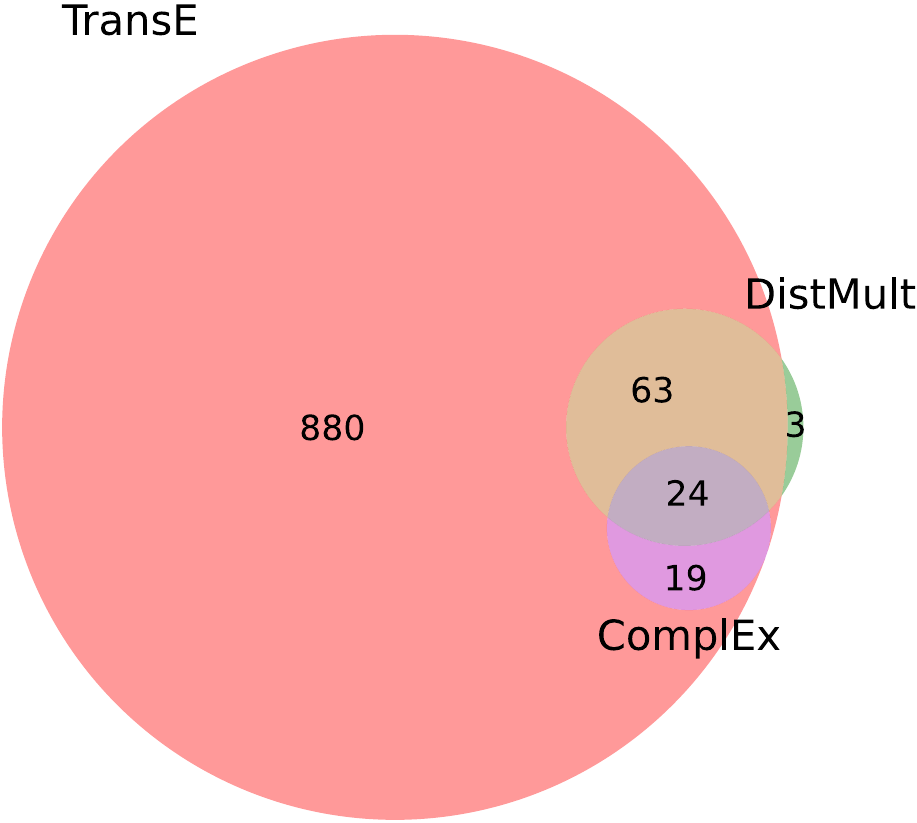}
    \caption{Family KG}
\end{subfigure}
\caption{Venn diagrams indicating the number of rules learned per method and
overlap for each KG.\@ Rules mined from different extensions count only once}%
\label{fig:overlap}
\end{figure}

\Cref{fig:overlap} shows how much the new rules extracted per KG embedding method overlap.
As expected, the majority of the rules derived from TransE-extended KGs are unique.
Also, every rule found with ComplEx was also found with TransE. Surprisingly, the overlap between DistMult and ComplEx was relatively small despite the two methods being similar.
In fact, every rule that was found both in ComplEx and DistMult-extended KGs was also found in KGs extended with TransE.
Besides, only DistMult-extended KGs generated rules that were not found in any TransE-extended KG.\@

As the random baseline method did not find new rules, we can conclude that the KG embedding methods identified non-trivial patterns in the data.
In \cref{fig:overlap}, we can observe that the choice of model can have a significant impact in the rules learned if we complete a KG using the embedding scores.
Furthermore, the results also highlight how limitations of the embedding models may impact the resulting rules.
However, we remark that the performance of the methods on KG completion tasks (discussed in \cref{sec:data_model_selection}) do not work well as predictors of the number or PCA confidence of the rules learned.

\subsubsection{Effect of Entity Selection Method and Rank Cutoff\label{sec:otherparams}}

Next, we discuss the impact of the other two parameters, entity selection strategy and rank cutoff value, on the quantity and PCA confidence of the rules mined by AMIE3.
In our analysis, we exclude rules that were only discovered by TransE, due to the large number of rules with low PCA confidence and the other issues due to TransE's ability to capture the semantics of the relations in  KGs, as we discussed before.

The \emph{probabilistic}, \emph{random} and \emph{least frequent} entity selection strategies have similar performance regarding PCA confidence (see \cref{fig:PCA_entity_boxplot}).
In comparison, the \emph{most frequent} strategy performs significantly worse.
However, in terms of number of rules learned, the \emph{most frequent} approach stands out as the best one.
This effect was more noticeable in the Family KG, in which AMIE3 mined a total of 105 new rules in KGs extended using the \emph{most frequent} strategy, while it mined only 36 new rules considering the second best, the \emph{random} approach.
We will investigate the causes of this behaviour in future works.
In the WN18RR KG all strategies allowed AMIE3 extract the original rules, but in the Family KG AMIE3 could find all original rules only when in KGs extended with the \emph{least frequent} method (see \cref{tab:table_rules_entities_family,tab:table_rules_entities_wn18rr} in the appendix).

The other relevant parameter is the rank cutoff value, which determines how many candidates were selected for each completion query.
Our initial assumption was that increasing the cutoff would worsen the overall PCA confidence of the new rules however it would increase the number of new rules learned.
Both in the Family and WN18RR KGs, we observed that the number of original rules extracted decreased as we increased the rank cutoff value, but the number of new rules mined increased, which is consistent with our assumption (see \cref{tab:table_rules_ranks_wn18rr,tab:table_rules_ranks_family} in the appendix).

However, the impact of the rank cutoff on the distribution of PCA rules was not as consistent as we expected (see \cref{fig:PCA_rank_boxplot} in the appendix).
In our experiments, the PCA confidence on the original data had similar medians on the three cutoff values in the WN18RR KG, but the proportion with low PCA increased when going from rank 1 to rank 4, and when going from rank 4 to rank 7 it was the proportion of rules with high PCA that increased instead.
When evaluating each rule on its source extension in the WN18RR case, the median of PCA confidence was significantly lower when the cutoff was rank 1, and even worse when the cutoff was 4.
Yet, the whole distribution of PCA confidence improved considerably when increasing the rank to 7.
One possible explanation for this behaviour is that with large cutoffs the new data ends up dominating the dataset, therefore, biasing the rules significantly.
In the case of the Family KG, increasing the rank cutoff improved the PCA confidence in general on the original dataset.
In fact, the whole distribution improves, including the median w.r.t.\ the median of the original rules.
On the extended data for the Family KG, we observed a slight improvement from rank 1 to rank 4, and rank 7 performed slightly worse than the others.
Therefore, it seems that the effects of rank cutoff depends more on the KG at hand.

\subsubsection{Rule Comparison}
We now discuss some of the rules mined by AMIE3 and their PCA confidence on the original KGs and extensions.
\Cref{wn18rr_original_rules_table_PCA} displays the original rules extracted by AMIE3 from the original WN18RR KG.\@
The \lcnamecref{wn18rr_original_rules_table_PCA} indicates in how many of the 48 
possible extensions (one for each combination of parameters)  each rule was mined (in the column ``Frequency'') and its PCA Confidence in the original KG.\@
Due to space constraints, we placed an analogous \lcnamecref{family_original_rules_table_PCA} for the Family KG in the appendix (\cref{family_original_rules_table_PCA}).

\begin{center}
\begin{table}[htbp]
\begin{tabular}{lrr}
\toprule
                                                                                                      Rule &  Frequency &  PCA Confidence \\
\midrule
                            DRF(y, x)   $\Rightarrow$ DRF(x, y) &           48 &        1.000000 \\
          IH(x, z) $\wedge$ SDTO(z, y)   $\Rightarrow$ SDTO(x, y) &           24 &        0.886525 \\
                   hypernym(x, z) $\wedge$ SDTO(z, y)   $\Rightarrow$ SDTO(x, y) &           36 &        0.828871 \\
                   has\_part(x, z) $\wedge$ SDTO(z, y)   $\Rightarrow$ SDTO(x, y) &           24 &        0.809524 \\
                   has\_part(z, x) $\wedge$ SDTO(z, y)   $\Rightarrow$ SDTO(x, y) &           24 &        0.771739 \\
                   hypernym(z, x) $\wedge$ SDTO(z, y)   $\Rightarrow$ SDTO(x, y) &           34 &        0.726667 \\
                                      has\_part(x, z) $\wedge$ IH(y, z)   $\Rightarrow$ has\_part(x, y) &           19 &        0.532544 \\
DRF(x, z) $\wedge$ SDTO(z, y)   $\Rightarrow$ SDTO(x, y) &           24 &        0.492857 \\
DRF(z, x) $\wedge$ SDTO(z, y)   $\Rightarrow$ SDTO(x, y) &           24 &        0.492857 \\
                                               has\_part(x, z) $\wedge$ hypernym(y, z)   $\Rightarrow$ has\_part(x, y) &           21 &        0.104016 \\
\bottomrule
\end{tabular}
\caption[Rules mined from the original WN18RR KG]{Rules mined from the original WN18RR KG, with their corresponding PCA confidences and how many times they were mined from the 48 extensions. DRF = ``\textit{derivationally related form}", STDO = ``\textit{synset domain topic of}" and IH = ``\textit{instance hypernym}".}%
\label{wn18rr_original_rules_table_PCA}
\end{table}
\end{center}

In both KGs, there was a strong correlation between the PCA confidence in the original dataset and the frequency a rule in it was mined from the extensions.
Using Spearmann's rank correlation, we get a coefficient of 0.65 (with p-value 0.04) for the WN18RR KG and 0.86 (with p-value $5.85e^{-29}$) for the Family KG.
In the Family KG, the only rules which were not mined in all extensions were those that use the predicate \(relative\).
There are two main possible causes: 
(1) this predicate is the least frequent relation in the Family KG and
(2)   the nature of the ``relative'' relation, which not only is symmetric and transitive, but also a 1-N relation that subsumes many of the other relations in the dataset, such as \(father\), \(mother\), and \(sibling\).



\section{Conclusion}\label{sec:conclusion}

We investigated the influence of distinct KGEs for KG completion by using rule mining to identify hidden patterns in the original KGs and after completion with a given KGE.
In most cases, no rules were lost but several new rules were added for the TransE approach. 
We conclude that the choice of KGE model strongly influences the number and quality of the rules mined.
Interestingly, the results for Link Prediction do not necessarily reflect on the confidence of the new rules obtained from extended KGs.
However, the PCA confidence of a rule in the original data can indicate  how likely it is to be mined in an extended (or ``completed'') KG.\@
Moreover, even models that are similar in formalisation and results in KG completion tasks (DistMult and ComplEx in our case) may differ considerably in the rules that they enable.
As future work, we mention an extension of this study with large KGs and fewer limitations on the number of candidates generated.

\newpage
\section*{Acknowledgements} The first author is supported by the ERC project “Lossy Preprocessing”
(LOPRE), grant number 819416, led by Prof. Saket Saurabh.
The second author is supported by the NFR project “Learning Description Logic
Ontologies”, grant number 316022.
%
\bibliography{bib_nnmdl}

\newpage
\appendix
\renewcommand\thefigure{\thesection.\arabic{figure}} 
\setcounter{figure}{0}

\renewcommand\thetable{\thesection.\arabic{table}} 
\setcounter{table}{0}

\section{Definition of PCA Confidence}\label{ap:pca}

For convenience of the reader, we provide 
the formal definition of PCA confidence~\cite{Lajus2020}. 
 We model a \emph{knowledge base} (KB) $\mathcal{K}$ as a set of assertions $r(s, o)$,
also called \emph{facts}, where $r$ is
a relation and $s,o$ (subject, object) are constants. 
A (Horn) \emph{rule} is an expression of the form $\mathbf{B}\Rightarrow H$ 
where $\mathbf{B}$ is a conjunction of atoms and $H$ is an atom (called \emph{head} of the rule). As already mentioned, a
rule is closed  if all variables appear in at least two atoms. 
 A substitution $\sigma$ is a partial mapping from variables to constants.
Substitutions can be straightforwardly extended to atoms and conjunctions.
Given a rule $R = B_1 \wedge \ldots \wedge B_n \Rightarrow H$ and a substitution $\sigma$, we
call $\sigma(R)$ an \emph{instantiation} of $R$. If $\sigma(B_i) \in \mathcal{K} \ \forall i \in \{1, \ldots, n\}$, we call $\sigma(H)$ a
prediction of $R$ from $\mathcal{K}$, and we write $\mathcal{K} \wedge R\models  \sigma(H)$. If
$ \sigma(H) \in \mathcal{K}$, we call $\sigma(H)$
a \emph{true prediction}.
The
\emph{functionality score}  of a relation $r$
is
\begin{equation} 
fun(r) = \frac{|\{s : \exists o :  r(s, o) \in\mathcal{K}\}|}
{|\{(s, o) : r(s, o) \in\mathcal{K}\}|}
\end{equation} 
If we have $r(s, o)$ in the KB $\mathcal{K}$, and
if $fun(r) \geq fun(r^{-})$, then we assume that all $r(s, o') \in\mathcal{K}$ do not hold in the real
world. If $fun(r) < fun(r^{-})$, then the PCA says that all $r(s, o') \in\mathcal{K}$ do not hold in
the real world. 
These assertions are 
\emph{false predictions}.  
The \emph{support of a rule $R$ in a KB $\mathcal{K}$} is the number
of true predictions $p$ (of the form $r(x,y )$) that the rule makes in the KB:\@
\begin{equation}
support(R) = |\{p : (\mathcal{K} \wedge R \models p) \wedge p \in \mathcal{K}\}| 
\end{equation}

The \emph{Partial Completeness Assumption} can be defined as follows.  
If we have $r(s, o)$ in the KB $\mathcal{K}$, and
if $fun(r) \geq fun(r^{-})$, then we assume that all 
$r(s, o') \in\mathcal{K}$
  do not hold in the real
world. If $fun(r) < fun(r^{-})$, then the PCA says that all $r(s', o) \in\mathcal{K}$ do not hold in
the real world.
The formula for \emph{PCA confidence} is 
\begin{equation}
pca\text{-}conf(\mathbf{B} \Rightarrow r(x, y)) = \frac{support(\mathbf{B} \Rightarrow r(x, y))}
{|\{(x, y) : \exists z: \mathbf{B} \wedge r(x, z)\}|}
\end{equation}
  for the case where $fun(r) \geq fun(r^{-})$. If $fun(r) < fun(r^{-})$, the denominator
becomes $|\{(x, y) : \exists x : B \wedge r(x, y)\}|$.

\section{Additional Data on Model Selection}

\Cref{tab:model_selection_space} displays the ranges of values given to the hyperparameters during model selection.

\begin{table}[htbp]
\centering
\begin{tabular}{@{}ll@{}}
\toprule
Hyperparameter                & Values                            \\ \midrule
Batches count                 & 50, 100                           \\
Epochs                        & 50, 100                           \\
Dimensions                    & 50, 100, 200                      \\
Negative samples per positive & 5, 10, 15                         \\
Loss function                 & pairwise, negative log likelihood \\
Pairwise margin loss          & 0.5, 1, 2                         \\ \bottomrule
\end{tabular}
\caption{Hyperparameter values selected for model selection}%
\label{tab:model_selection_space}
\end{table}

\section{Additional Data for \Cref{sec:results}}

This section includes additional tables and plots that support the discussion in \Cref{sec:results}.

\begin{longtable}{rrrrrr}
    \toprule
\textbf{child} & \textbf{father} & \textbf{mother} & \textbf{relative} & \textbf{sibling} & \textbf{spouse} \\ \midrule
0.009  & -0.010 & -0.010 & -0.001   & 0.000   & 0.000  \\
0.002  & -0.002 & -0.002 & 0.000    & 0.000   & 0.000  \\
-0.001 & 0.001  & 0.002  & 0.001    & 0.000   & 0.000  \\
-0.005 & 0.005  & 0.005  & 0.000    & 0.000   & 0.000  \\
-0.012 & 0.012  & 0.012  & 0.001    & 0.000   & 0.001  \\
-0.007 & 0.008  & 0.007  & 0.001    & 0.000   & -0.001 \\
0.006  & -0.006 & -0.006 & -0.001   & 0.000   & 0.000  \\
-0.004 & 0.005  & 0.004  & 0.001    & 0.000   & 0.000  \\
0.003  & -0.002 & -0.002 & 0.001    & 0.000   & 0.000  \\
0.016  & -0.016 & -0.015 & 0.000    & 0.000   & 0.000  \\
-0.002 & 0.002  & 0.002  & 0.000    & 0.000   & 0.000  \\
-0.001 & 0.001  & 0.001  & 0.001    & 0.000   & 0.000  \\
0.006  & -0.005 & -0.006 & -0.001   & 0.000   & 0.000  \\
0.003  & -0.004 & -0.004 & 0.000    & 0.000   & 0.000  \\
-0.001 & 0.001  & 0.002  & -0.001   & 0.000   & 0.000  \\
-0.013 & 0.014  & 0.013  & -0.001   & 0.000   & 0.000  \\
0.008  & -0.008 & -0.008 & 0.000    & 0.000   & 0.000  \\
0.006  & -0.006 & -0.006 & -0.002   & 0.000   & 0.000  \\
-0.013 & 0.013  & 0.013  & 0.000    & 0.000   & 0.000  \\
-0.009 & 0.009  & 0.009  & 0.000    & 0.000   & 0.000  \\
-0.005 & 0.005  & 0.005  & 0.001    & 0.000   & 0.000  \\
-0.007 & 0.007  & 0.007  & 0.001    & 0.000   & 0.000  \\
0.001  & -0.001 & -0.001 & 0.000    & 0.000   & 0.001  \\
0.009  & -0.009 & -0.008 & -0.001   & 0.000   & 0.000  \\
-0.003 & 0.003  & 0.003  & 0.000    & 0.000   & 0.000  \\
0.003  & -0.003 & -0.003 & 0.000    & 0.000   & 0.000  \\
-0.014 & 0.014  & 0.014  & 0.001    & 0.000   & 0.001  \\
0.005  & -0.006 & -0.005 & 0.000    & 0.000   & 0.001  \\
-0.006 & 0.005  & 0.005  & 0.000    & 0.000   & 0.001  \\
0.000  & 0.000  & 0.000  & 0.001    & 0.000   & 0.000  \\
-0.003 & 0.003  & 0.003  & 0.000    & 0.000   & 0.000  \\
-0.003 & 0.003  & 0.004  & 0.001    & 0.000   & 0.000  \\
-0.009 & 0.009  & 0.009  & -0.001   & 0.000   & 0.000  \\
-0.011 & 0.011  & 0.010  & 0.001    & 0.000   & 0.000  \\
-0.001 & 0.001  & 0.001  & -0.001   & 0.000   & 0.000  \\
-0.009 & 0.009  & 0.008  & -0.001   & 0.000   & -0.001 \\
0.006  & -0.006 & -0.006 & -0.001   & 0.000   & 0.000  \\
-0.002 & 0.002  & 0.002  & 0.002    & 0.000   & 0.000  \\
0.002  & -0.002 & -0.002 & 0.000    & 0.000   & 0.000  \\
0.005  & -0.005 & -0.005 & -0.002   & 0.000   & 0.000  \\
-0.005 & 0.005  & 0.004  & 0.001    & 0.000   & 0.000  \\
0.007  & -0.007 & -0.006 & 0.001    & 0.000   & 0.000  \\
0.004  & -0.004 & -0.003 & 0.000    & 0.000   & 0.000  \\
0.005  & -0.006 & -0.006 & 0.001    & 0.000   & 0.000  \\
-0.005 & 0.006  & 0.005  & 0.001    & 0.000   & 0.001  \\
0.007  & -0.008 & -0.008 & -0.001   & 0.000   & 0.000  \\
0.005  & -0.005 & -0.004 & 0.000    & 0.000   & -0.001 \\
0.002  & -0.002 & -0.002 & -0.001   & 0.000   & 0.000  \\
0.003  & -0.003 & -0.002 & -0.001   & 0.000   & 0.001  \\
0.007  & -0.007 & -0.007 & 0.001    & 0.000   & 0.000  \\ \bottomrule
\caption{TransE's 50-dimensional embedding vectors for the different relations in the family dataset, rounded to the third decimal. This table shows that the vectors for all relations are close to the zero-vector.}%
\label{TransE_embedding_family}
\end{longtable}

\begin{longtable}{rrrrrr}
\toprule
\textbf{DRF} & \textbf{has\_part} & \textbf{hypernym} & \textbf{IH} & \textbf{MM} & \textbf{SDT} \\ \midrule
0.000        & -0.002             & -0.003            & -0.004      & -0.027      & -0.028       \\
0.000        & 0.000              & 0.012             & 0.081       & -0.001      & 0.010        \\
0.000        & -0.003             & 0.007             & 0.010       & -0.012      & 0.009        \\
0.000        & -0.003             & 0.003             & -0.148      & -0.042      & 0.000        \\
0.000        & 0.017              & 0.003             & 0.005       & 0.002       & 0.007        \\
0.000        & -0.001             & 0.002             & 0.139       & -0.014      & 0.003        \\
0.000        & -0.002             & -0.002            & -0.115      & 0.031       & -0.006       \\
0.000        & -0.006             & 0.005             & 0.005       & -0.015      & 0.008        \\
0.000        & -0.001             & 0.001             & 0.150       & -0.021      & 0.010        \\
0.000        & -0.007             & -0.011            & 0.119       & 0.012       & 0.019        \\
0.000        & -0.004             & 0.013             & 0.118       & -0.019      & 0.012        \\
0.000        & -0.004             & -0.002            & -0.069      & 0.042       & -0.001       \\
0.000        & -0.002             & 0.002             & -0.165      & -0.025      & 0.002        \\
0.000        & 0.000              & -0.012            & -0.017      & 0.007       & -0.012       \\
0.000        & -0.018             & -0.002            & -0.111      & 0.008       & 0.015        \\
0.000        & 0.021              & 0.001             & 0.006       & 0.003       & -0.041       \\
0.000        & 0.001              & -0.015            & 0.005       & 0.043       & -0.026       \\
0.000        & -0.002             & 0.005             & -0.152      & -0.017      & -0.006       \\
0.000        & 0.002              & 0.002             & -0.082      & -0.002      & -0.009       \\
0.000        & -0.002             & 0.001             & 0.134       & -0.048      & 0.044        \\
0.000        & -0.001             & 0.007             & 0.142       & -0.003      & 0.016        \\
0.000        & -0.001             & -0.001            & -0.134      & -0.060      & -0.009       \\
0.000        & 0.018              & -0.003            & -0.142      & -0.020      & -0.016       \\
0.000        & 0.002              & -0.008            & -0.084      & 0.010       & -0.030       \\
0.000        & -0.004             & 0.009             & -0.062      & -0.007      & -0.019       \\
0.000        & 0.005              & 0.005             & 0.067       & 0.000       & 0.011        \\
0.000        & -0.003             & 0.005             & 0.113       & -0.009      & 0.028        \\
0.000        & 0.002              & -0.005            & 0.002       & -0.038      & -0.010       \\
0.000        & 0.007              & 0.001             & -0.127      & 0.000       & -0.010       \\
0.000        & -0.002             & 0.000             & 0.036       & -0.007      & 0.013        \\
0.001        & -0.005             & 0.002             & 0.002       & 0.009       & -0.010       \\
0.000        & -0.004             & 0.004             & -0.075      & -0.045      & 0.017        \\
-0.001       & 0.016              & 0.001             & 0.110       & 0.018       & -0.024       \\
0.000        & 0.009              & 0.000             & 0.156       & -0.002      & -0.020       \\
0.001        & 0.000              & 0.007             & 0.023       & -0.043      & -0.001       \\
0.000        & -0.002             & -0.003            & -0.103      & 0.001       & 0.000        \\
0.000        & -0.008             & 0.007             & 0.167       & -0.005      & 0.019        \\
0.000        & 0.000              & -0.017            & 0.021       & 0.012       & -0.020       \\
0.000        & -0.007             & -0.003            & -0.001      & 0.025       & -0.002       \\
0.000        & -0.008             & -0.006            & 0.029       & 0.047       & 0.006        \\
-0.001       & 0.002              & 0.000             & -0.001      & 0.047       & -0.004       \\
0.000        & -0.002             & 0.014             & 0.139       & -0.036      & 0.005        \\
0.000        & 0.001              & 0.010             & 0.010       & -0.056      & 0.003        \\
0.000        & -0.001             & 0.001             & -0.005      & 0.000       & 0.019        \\
0.000        & -0.002             & -0.001            & -0.128      & 0.005       & 0.007        \\
0.000        & -0.001             & 0.002             & 0.131       & 0.001       & -0.002       \\
0.000        & -0.004             & 0.011             & 0.137       & -0.008      & 0.024        \\
0.000        & 0.005              & -0.002            & -0.092      & -0.045      & -0.006       \\
0.000        & 0.001              & 0.011             & -0.058      & -0.005      & -0.023       \\
0.000        & 0.004              & -0.013            & -0.041      & 0.021       & -0.046      \\ \bottomrule
\caption{TransE's 50-dimensional embedding vectors for the different relations in the WN18RR dataset, rounded to the third decimal. This table shows that the vectors for all relations are close to the zero-vector. \textbf{DRF} = \textit{derivationally\_related\_form}, \textbf{IH} = \textit{instance\_hypernym}, \textbf{MM} = \textit{member\_meronym} and \textbf{SDT} = \textit{synset\_domain\_topic of}.}%
\label{TransE_embedding_WN18RR}
\end{longtable}

\begin{figure}[htbp]
\centering
\begin{subfigure}{.5\textwidth}
  \centering
  \includegraphics[width=1\linewidth]{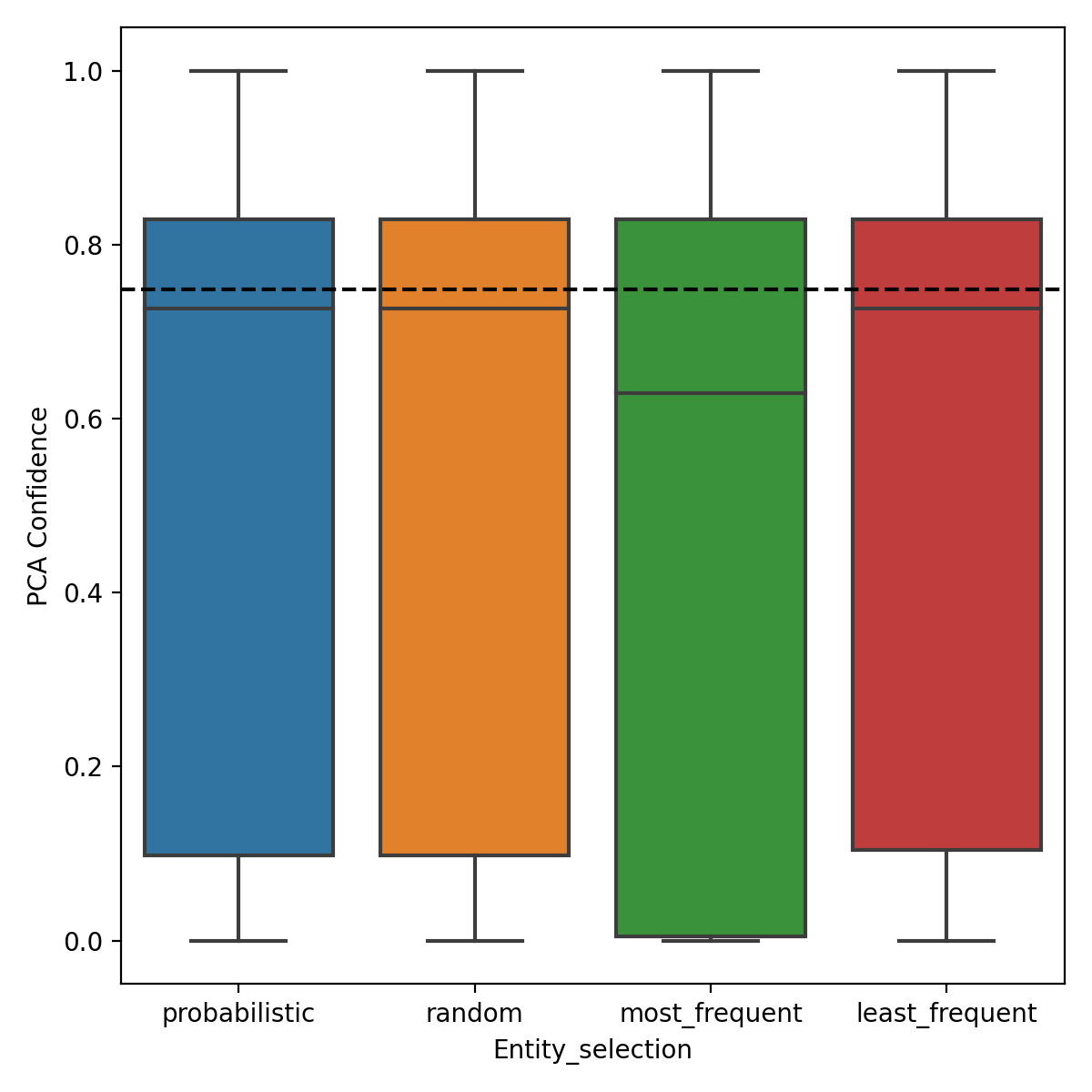}
  \caption{Original WN18RR KG}%
  \label{fig:PCA-entity_wn18rr_boxplot_sub}
\end{subfigure}%
\begin{subfigure}{.5\textwidth}
  \centering
  \includegraphics[width=1\linewidth]{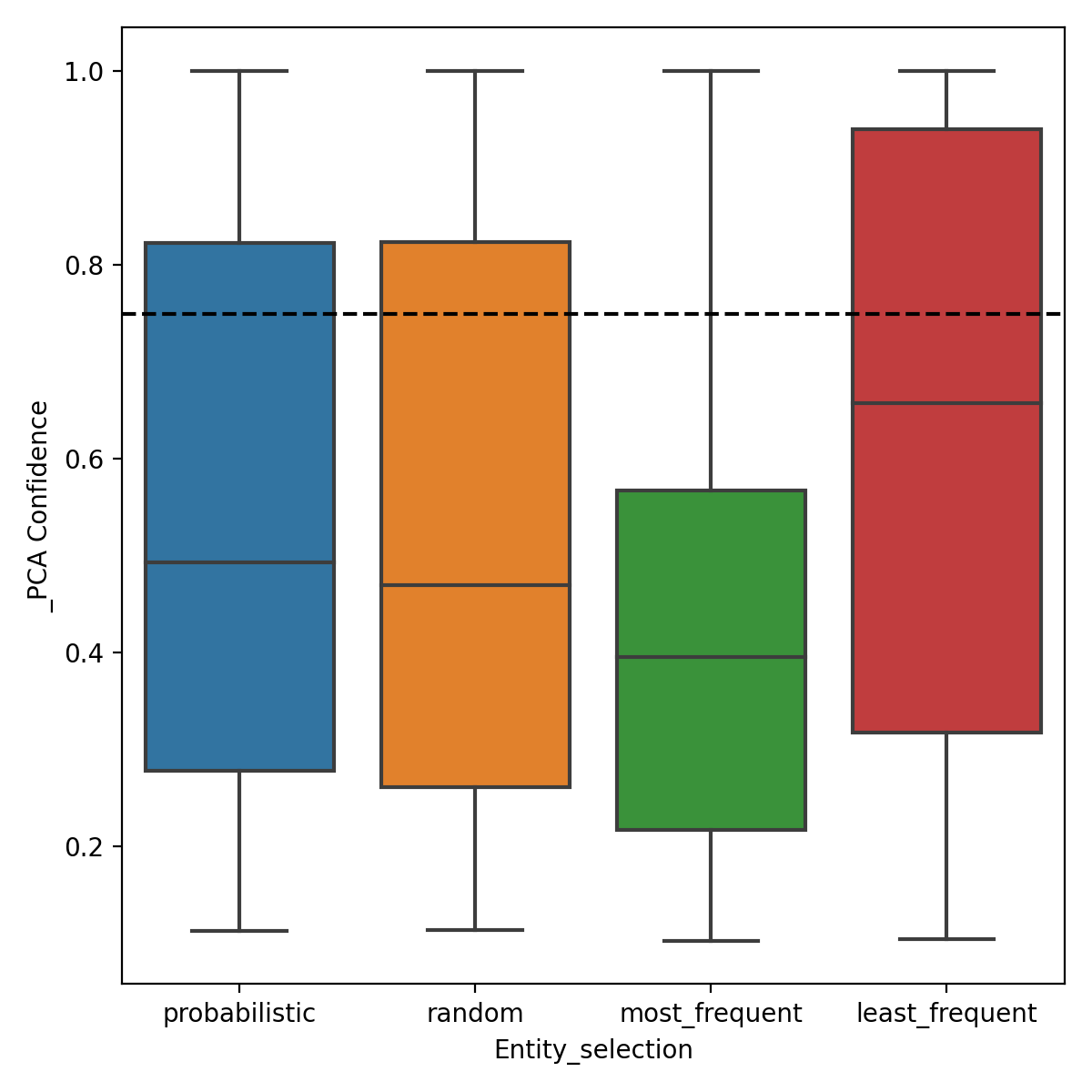}
  \caption{Extended WN18RR KG}%
  \label{fig:_PCA_entity_wn18rr_boxplot_sub}
\end{subfigure}
\begin{subfigure}{.5\textwidth}
  \centering
  \includegraphics[width=1\linewidth]{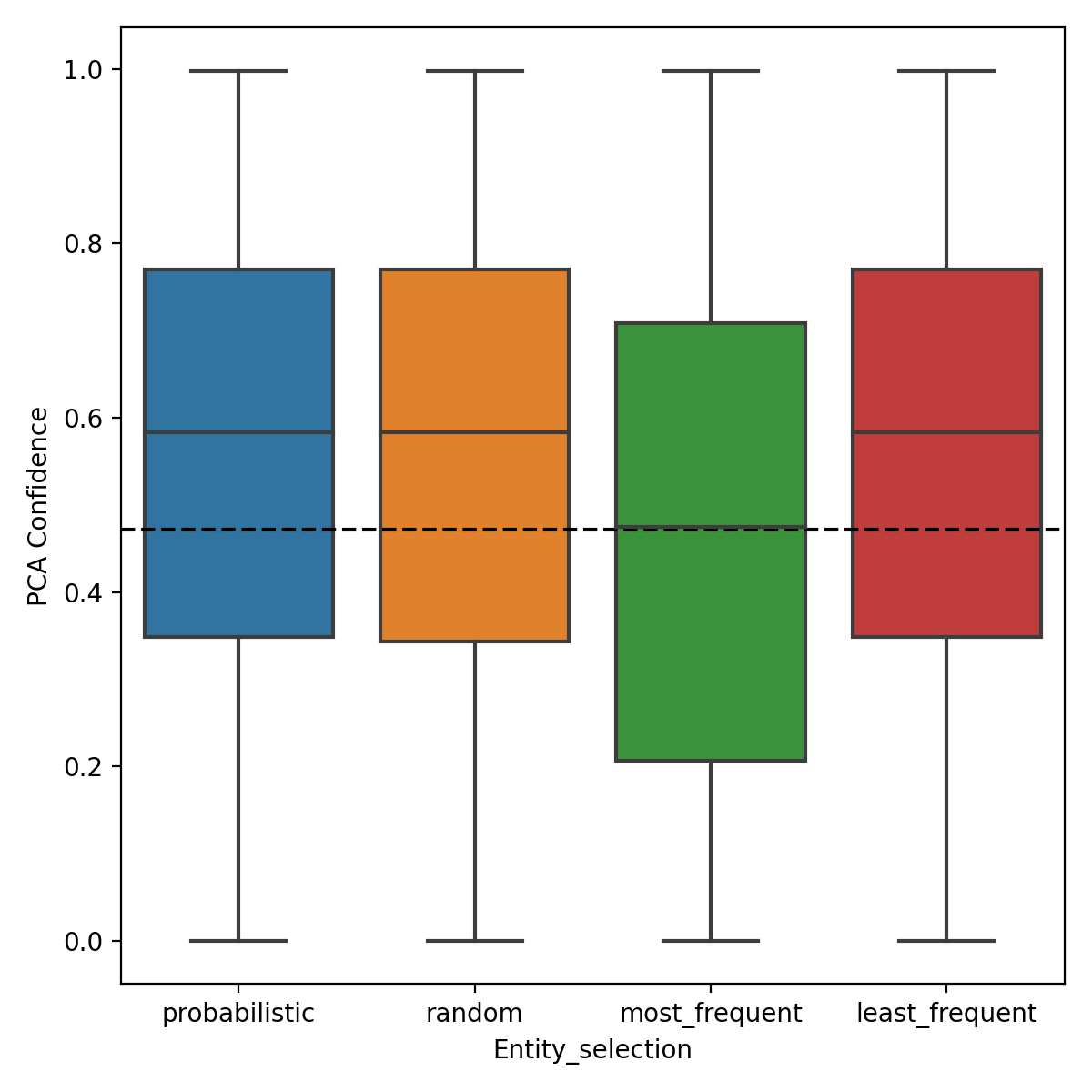}
  \caption{Original family KG}%
  \label{fig:models_entity_boxplot_sub}
\end{subfigure}%
\begin{subfigure}{.5\textwidth}
  \centering
  \includegraphics[width=1\linewidth]{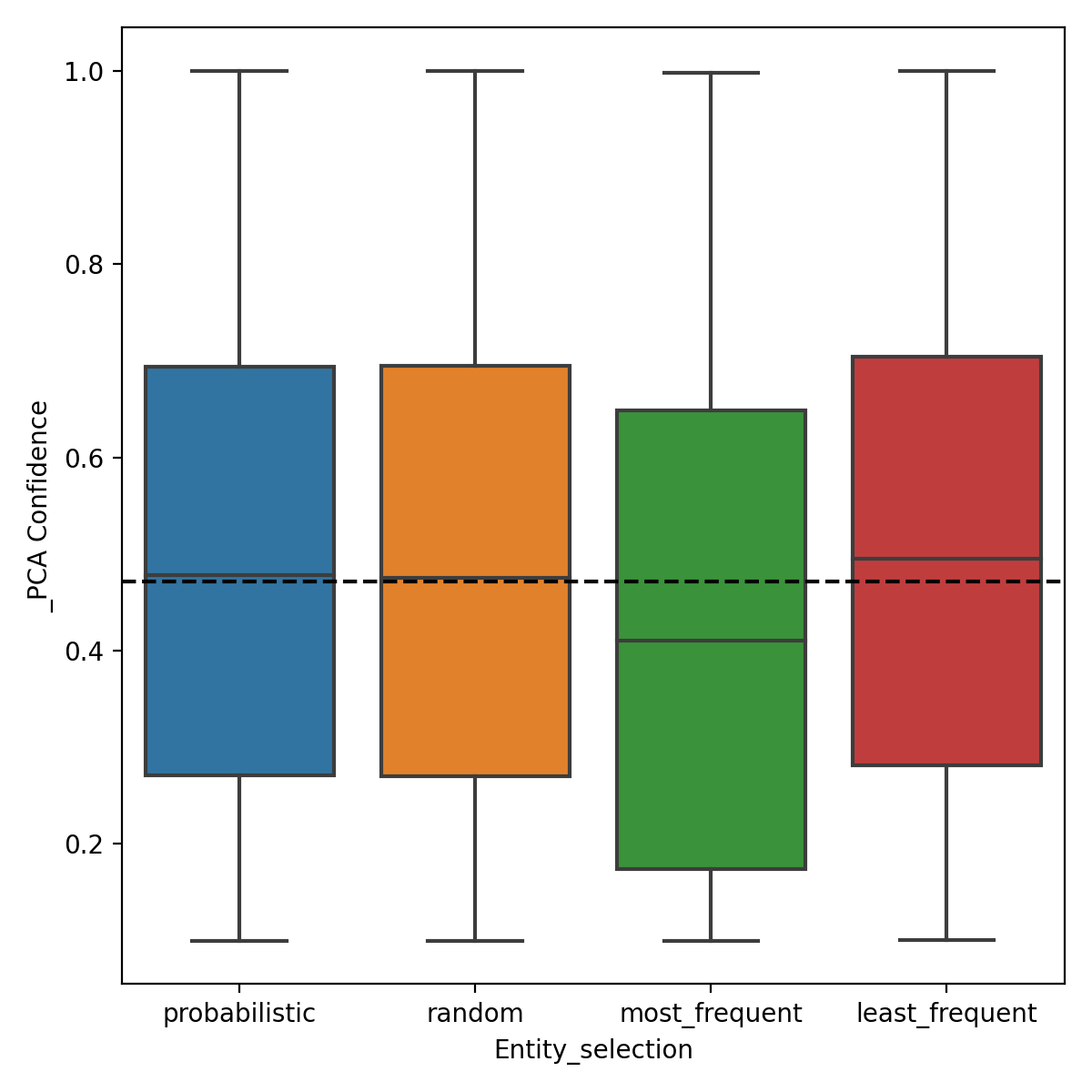}
  \caption{Extended family KG}%
  \label{fig:_PCA_entity_family_boxplot_sub}
\end{subfigure}
\caption[Dist.\ of PCA conf.\ of rules by entity selection strategies]{Distribution of PCA confidence of mined rules by entity selection strategy. PCA confidence scores are calculated on the original KG and the extended KGs from which the rules are mined. Rules may count multiple times if they were mined from different extensions. The dashed line represents the median PCA confidence of the rules mined from the original KG.}%
\label{fig:PCA_entity_boxplot}
\end{figure}

\begin{table}[]
\centering
\begin{tabular}{@{}lrrrrrr@{}}
\toprule
Strategy       & \multicolumn{2}{l}{Original}                                   & \multicolumn{2}{l}{New}                                        & \multicolumn{2}{l}{Missed}                                        \\ \midrule
               & \multicolumn{1}{c}{Quantity} & \multicolumn{1}{c}{\% of mined} & \multicolumn{1}{c}{Quantity} & \multicolumn{1}{c}{\% of mined} & \multicolumn{1}{c}{Quantity} & \multicolumn{1}{c}{\% of original} \\ \cmidrule(l){2-7} 
Random         & 10                           & 53\%                            & 9                            & 47\%                            & 0                            & 0\%                                \\
Most frequent  & 10                           & 42\%                            & 14                           & 58\%                            & 0                            & 0\%                                \\
Least frequent & 10                           & 48\%                            & 11                           & 52\%                            & 0                            & 0\%                                \\
Probabilistic  & 10                           & 45\%                            & 12                           & 55\%                            & 0                            & 0\%                                \\ \bottomrule
\end{tabular}
\caption{Count of rules mined or missed by entity selection strategy for the WN18RR KG.}%
\label{tab:table_rules_entities_wn18rr}
\end{table}

\begin{table}[]
\centering
\begin{tabular}{@{}lrrrrrr@{}}
\toprule
Strategy       & \multicolumn{2}{l}{Original}                                   & \multicolumn{2}{l}{New}                                        & \multicolumn{2}{l}{Missed}                                        \\ \midrule
               & \multicolumn{1}{c}{Quantity} & \multicolumn{1}{c}{\% of mined} & \multicolumn{1}{c}{Quantity} & \multicolumn{1}{c}{\% of mined} & \multicolumn{1}{c}{Quantity} & \multicolumn{1}{c}{\% of original} \\ \cmidrule(l){2-7} 
Random         & 90                           & 71\%                            & 36                           & 29\%                            & 4                            & 4\%                                \\
Most frequent  & 88                           & 46\%                            & 105                          & 54\%                            & 6                            & 6\%                                \\
Least frequent & 94                           & 77\%                            & 28                           & 33\%                            & 0                            & 0\%                                \\
Probabilistic  & 91                           & 73\%                            & 33                           & 27\%                            & 3                            & 3\%                                \\ \bottomrule
\end{tabular}
\caption{Count of rules mined or missed by entity selection strategy for the Family KG.}%
\label{tab:table_rules_entities_family}
\end{table}

\begin{table}[]
\centering
\begin{tabular}{@{}crrrrrr@{}}
\toprule
Rank cutoff & \multicolumn{2}{c}{Kept}                                       & \multicolumn{2}{c}{New}                                        & \multicolumn{2}{c}{Missed}                                        \\ \midrule
            & \multicolumn{1}{c}{Quantity} & \multicolumn{1}{c}{\% of mined} & \multicolumn{1}{c}{Quantity} & \multicolumn{1}{c}{\% of mined} & \multicolumn{1}{c}{Quantity} & \multicolumn{1}{c}{\% of original} \\ \cmidrule(l){2-7} 
1           & 10                           & 50\%                            & 10                           & 50\%                            & 0                            & 0\%                                \\
4           & 4                            & 27\%                            & 11                           & 73\%                            & 6                            & 60\%                               \\
7           & 3                            & 20\%                            & 12                           & 80\%                            & 7                            & 70\%                               \\ \bottomrule
\end{tabular}
\caption{Count of rules mined or missed by rank cutoff value for the WN18RR KG (excluding TransE-only rules).}%
\label{tab:table_rules_ranks_wn18rr}
\end{table}

\begin{table}[]
\centering
\begin{tabular}{@{}crrrrrr@{}}
\toprule
\multicolumn{1}{c}{Rank cutoff} & \multicolumn{2}{c}{Original}                                   & \multicolumn{2}{c}{New}                                        & \multicolumn{2}{c}{Missed}                                        \\ \midrule
                                & \multicolumn{1}{c}{Quantity} & \multicolumn{1}{c}{\% of mined} & \multicolumn{1}{c}{Quantity} & \multicolumn{1}{c}{\% of mined} & \multicolumn{1}{c}{Quantity} & \multicolumn{1}{c}{\% of original} \\ \cmidrule(l){2-7} 
1                               & 94                           & 94\%                            & 11                           & 10\%                            & 0                            & 0\%                                \\
4                               & 90                           & 51\%                            & 85                           & 49\%                            & 4                            & 4\%                                \\
7                               & 88                           & 45\%                            & 108                          & 55\%                            & 6                            & 6\%                                \\ \bottomrule
\end{tabular}
\caption{Count of rules mined or missed by rank cutoff value for the Family KG (excluding TransE-only rules).}%
\label{tab:table_rules_ranks_family}
\end{table}

\begin{figure}[!htb]
\centering
\begin{subfigure}{.5\textwidth}
  \centering
  \includegraphics[width=1\linewidth]{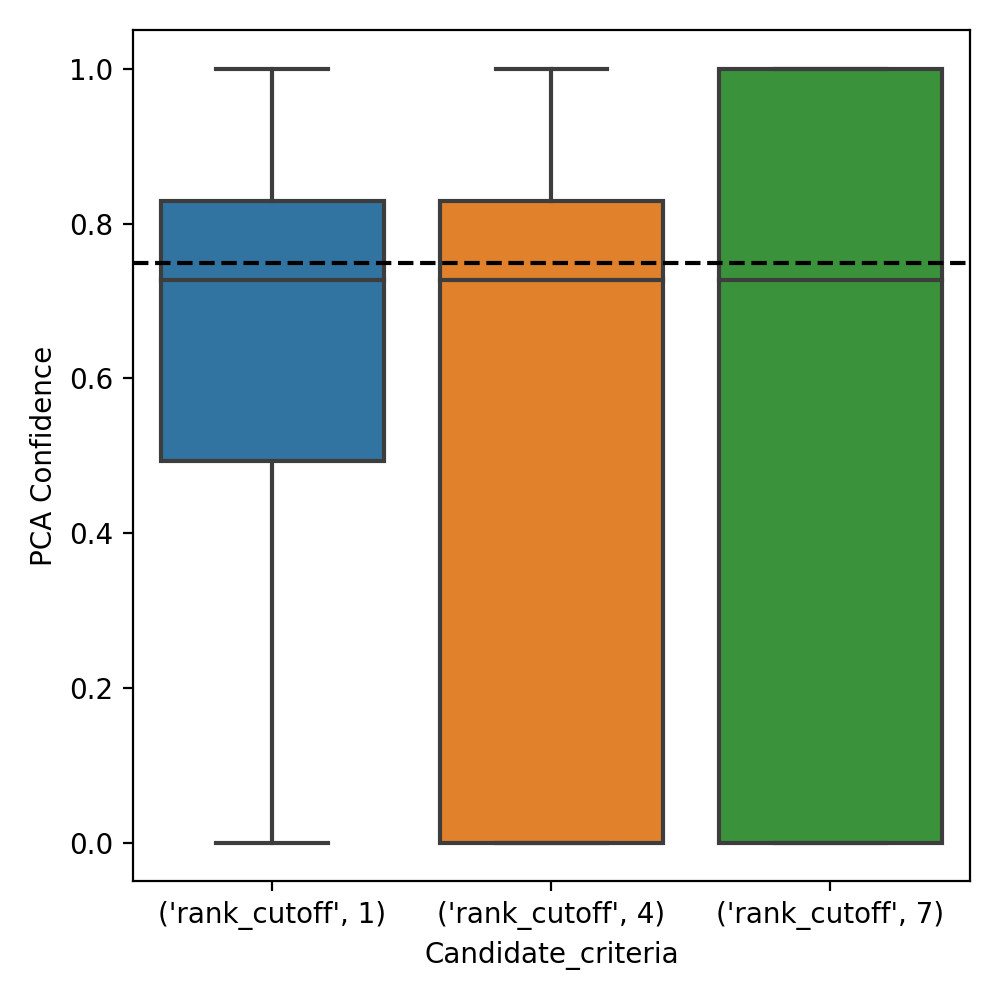}
  \caption{Original WN18RR KG}%
  \label{fig:PCA-rank_wn18rr_boxplot_sub}
\end{subfigure}%
\begin{subfigure}{.5\textwidth}
  \centering
  \includegraphics[width=1\linewidth]{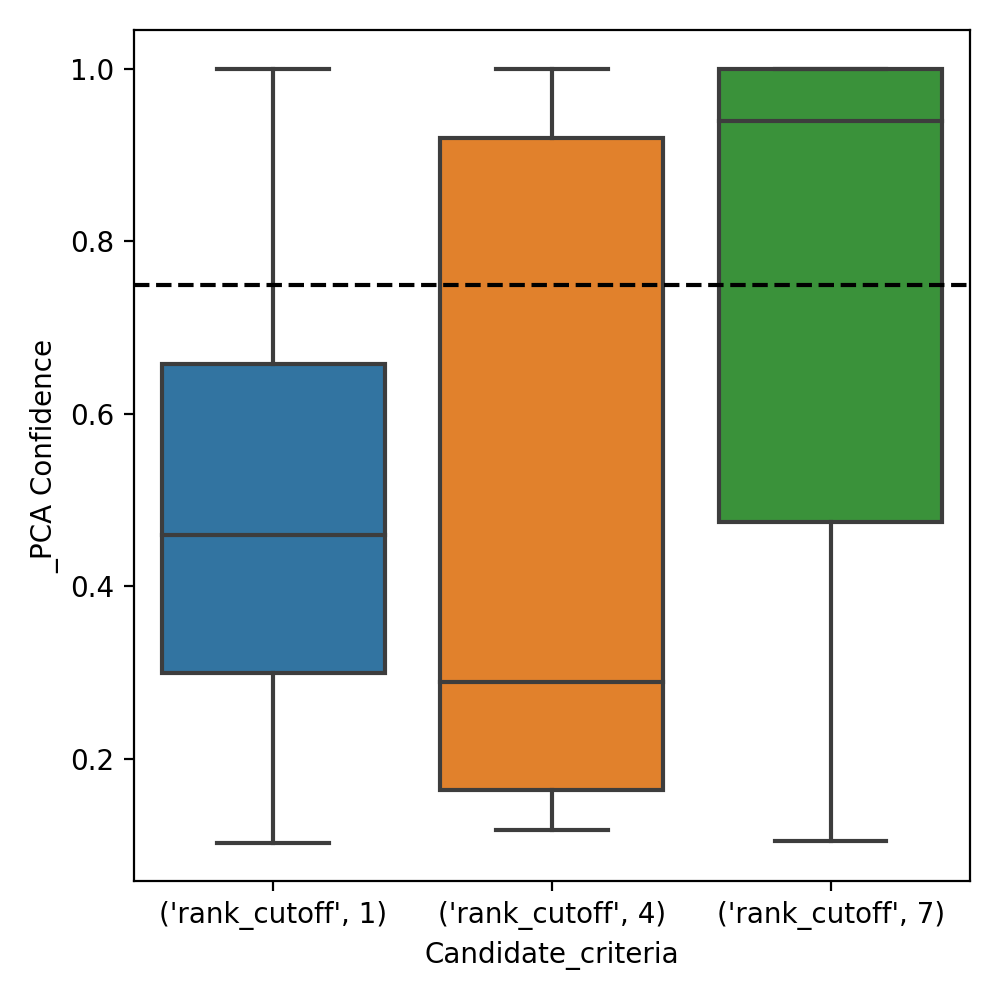}
  \caption{Extended WN18RR KG}%
  \label{fig:_PCA_rank_wn18rr_boxplot_sub}
\end{subfigure}
\begin{subfigure}{.5\textwidth}
  \centering
  \includegraphics[width=1\linewidth]{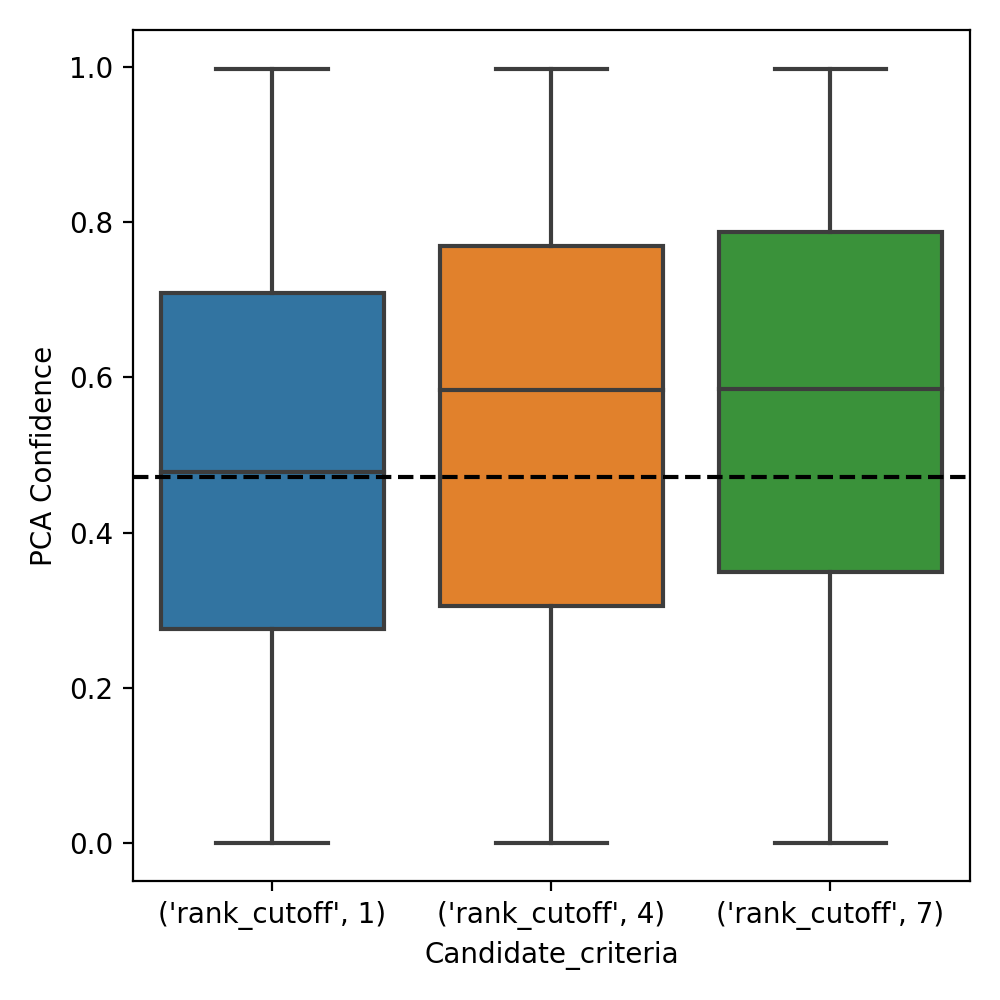}
  \caption{Original family KG}%
  \label{fig:models_rank_boxplot_sub}
\end{subfigure}%
\begin{subfigure}{.5\textwidth}
  \centering
  \includegraphics[width=1\linewidth]{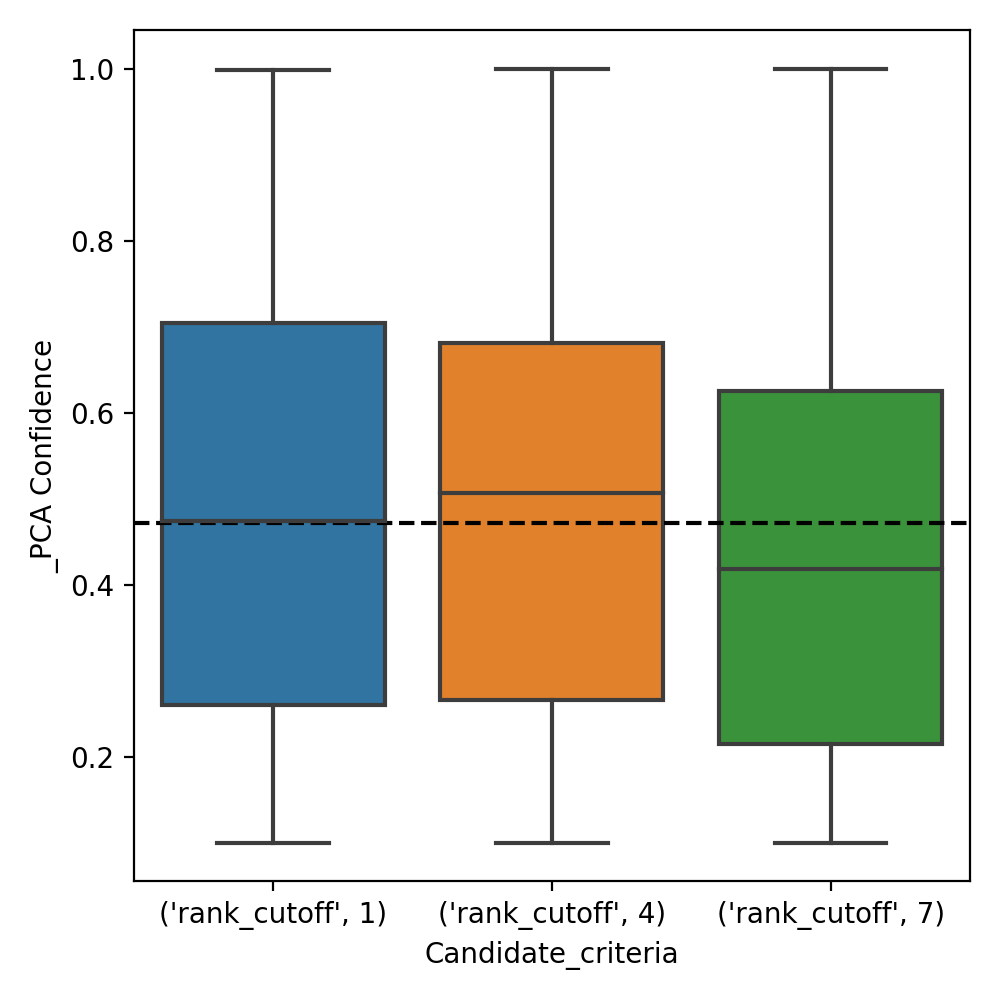}
  \caption{Extended family KG}%
  \label{fig:_PCA_rank_family_boxplot_sub}
\end{subfigure}
\caption[Dist.\ of PCA conf.\ of rules by rank cutoff values]{Distribution of PCA confidence of mined rules by rank cutoff values. PCA confidence scores are calculated on the original KG and the extended KGs from which the rules are mined. Rules may count multiple times if they were mined from different extensions. The dashed line represents the median PCA confidence of the rules mined from the original KG.}%
\label{fig:PCA_rank_boxplot}
\end{figure}

\begin{longtable}{lrr}
\toprule
                                                    Rule &  Frequency &  PCA Confidence \\
\midrule
                       father(y, x)   $\Rightarrow$ child(x, y) &           48 &        0.998071 \\
                      spouse(y, x)   $\Rightarrow$ spouse(x, y) &           48 &        0.992355 \\
                       mother(y, x)   $\Rightarrow$ child(x, y) &           48 &        0.991906 \\
                    sibling(y, x)   $\Rightarrow$ sibling(x, y) &           48 &        0.989634 \\
      father(z, y) $\wedge$ mother(z, x)   $\Rightarrow$ spouse(x, y) &           48 &        0.967952 \\
     father(z, y) $\wedge$ sibling(z, x)   $\Rightarrow$ father(x, y) &           48 &        0.961199 \\
     father(z, y) $\wedge$ sibling(x, z)   $\Rightarrow$ father(x, y) &           48 &        0.961023 \\
      father(z, x) $\wedge$ sibling(y, z)   $\Rightarrow$ child(x, y) &           48 &        0.954603 \\
      father(z, x) $\wedge$ sibling(z, y)   $\Rightarrow$ child(x, y) &           48 &        0.954328 \\
      father(z, x) $\wedge$ mother(z, y)   $\Rightarrow$ spouse(x, y) &           48 &        0.953170 \\
      mother(x, z) $\wedge$ spouse(z, y)   $\Rightarrow$ father(x, y) &           48 &        0.872686 \\
      mother(x, z) $\wedge$ spouse(y, z)   $\Rightarrow$ father(x, y) &           48 &        0.867655 \\
       child(x, z) $\wedge$ sibling(y, z)   $\Rightarrow$ child(x, y) &           48 &        0.851012 \\
       child(x, z) $\wedge$ sibling(z, y)   $\Rightarrow$ child(x, y) &           48 &        0.850519 \\
                  relative(y, x)   $\Rightarrow$ relative(x, y) &           48 &        0.846306 \\
       mother(y, z) $\wedge$ spouse(z, x)   $\Rightarrow$ child(x, y) &           48 &        0.839947 \\
       mother(y, z) $\wedge$ spouse(x, z)   $\Rightarrow$ child(x, y) &           48 &        0.838288 \\
     mother(z, y) $\wedge$ sibling(x, z)   $\Rightarrow$ mother(x, y) &           48 &        0.787816 \\
     mother(z, y) $\wedge$ sibling(z, x)   $\Rightarrow$ mother(x, y) &           48 &        0.787401 \\
                       child(y, x)   $\Rightarrow$ father(x, y) &           48 &        0.769746 \\
      child(z, x) $\wedge$ mother(y, z)   $\Rightarrow$ sibling(x, y) &           48 &        0.709805 \\
      child(z, y) $\wedge$ mother(x, z)   $\Rightarrow$ sibling(x, y) &           48 &        0.709139 \\
      father(x, z) $\wedge$ spouse(z, y)   $\Rightarrow$ mother(x, y) &           48 &        0.708957 \\
     mother(x, z) $\wedge$ mother(y, z)   $\Rightarrow$ sibling(x, y) &           48 &        0.707149 \\
      mother(z, x) $\wedge$ sibling(y, z)   $\Rightarrow$ child(x, y) &           48 &        0.704016 \\
      mother(z, x) $\wedge$ sibling(z, y)   $\Rightarrow$ child(x, y) &           48 &        0.703385 \\
      father(x, z) $\wedge$ spouse(y, z)   $\Rightarrow$ mother(x, y) &           48 &        0.701812 \\
        child(z, y) $\wedge$ spouse(x, z)   $\Rightarrow$ child(x, y) &           48 &        0.688723 \\
        child(z, y) $\wedge$ spouse(z, x)   $\Rightarrow$ child(x, y) &           48 &        0.687653 \\
      child(z, x) $\wedge$ father(y, z)   $\Rightarrow$ sibling(x, y) &           48 &        0.668446 \\
       child(z, x) $\wedge$ child(z, y)   $\Rightarrow$ sibling(x, y) &           48 &        0.664745 \\
      child(z, y) $\wedge$ father(x, z)   $\Rightarrow$ sibling(x, y) &           48 &        0.662206 \\
     father(x, z) $\wedge$ father(y, z)   $\Rightarrow$ sibling(x, y) &           48 &        0.658865 \\
   sibling(x, z) $\wedge$ sibling(y, z)   $\Rightarrow$ sibling(x, y) &           48 &        0.590059 \\
       father(y, z) $\wedge$ spouse(z, x)   $\Rightarrow$ child(x, y) &           48 &        0.588533 \\
       father(y, z) $\wedge$ spouse(x, z)   $\Rightarrow$ child(x, y) &           48 &        0.588257 \\
   sibling(z, y) $\wedge$ sibling(x, z)   $\Rightarrow$ sibling(x, y) &           48 &        0.588151 \\
   sibling(z, x) $\wedge$ sibling(y, z)   $\Rightarrow$ sibling(x, y) &           48 &        0.584857 \\
      child(y, z) $\wedge$ sibling(x, z)   $\Rightarrow$ father(x, y) &           48 &        0.584069 \\
      child(y, z) $\wedge$ sibling(z, x)   $\Rightarrow$ father(x, y) &           48 &        0.583590 \\
   sibling(z, x) $\wedge$ sibling(z, y)   $\Rightarrow$ sibling(x, y) &           48 &        0.564657 \\
                       child(y, x)   $\Rightarrow$ mother(x, y) &           48 &        0.535350 \\
 relative(x, z) $\wedge$ sibling(y, z)   $\Rightarrow$ relative(x, y) &           40 &        0.484528 \\
 relative(x, z) $\wedge$ sibling(z, y)   $\Rightarrow$ relative(x, y) &           40 &        0.483459 \\
       child(y, z) $\wedge$ mother(z, x)   $\Rightarrow$ spouse(x, y) &           48 &        0.480628 \\
       child(x, z) $\wedge$ mother(z, y)   $\Rightarrow$ spouse(x, y) &           48 &        0.477785 \\
        child(x, z) $\wedge$ child(y, z)   $\Rightarrow$ spouse(x, y) &           48 &        0.471973 \\
       child(x, z) $\wedge$ father(z, y)   $\Rightarrow$ spouse(x, y) &           48 &        0.471666 \\
       child(y, z) $\wedge$ father(z, x)   $\Rightarrow$ spouse(x, y) &           48 &        0.469716 \\
      child(y, z) $\wedge$ sibling(x, z)   $\Rightarrow$ mother(x, y) &           48 &        0.422559 \\
      child(y, z) $\wedge$ sibling(z, x)   $\Rightarrow$ mother(x, y) &           48 &        0.422363 \\
       child(z, x) $\wedge$ spouse(z, y)   $\Rightarrow$ mother(x, y) &           48 &        0.386296 \\
       child(z, x) $\wedge$ spouse(y, z)   $\Rightarrow$ mother(x, y) &           48 &        0.382227 \\
  father(y, z) $\wedge$ relative(x, z)   $\Rightarrow$ relative(x, y) &           37 &        0.378085 \\
  father(z, y) $\wedge$ relative(x, z)   $\Rightarrow$ relative(x, y) &           37 &        0.372340 \\
   child(z, y) $\wedge$ relative(x, z)   $\Rightarrow$ relative(x, y) &           37 &        0.364472 \\
       child(z, x) $\wedge$ spouse(y, z)   $\Rightarrow$ father(x, y) &           48 &        0.351479 \\
 relative(z, x) $\wedge$ sibling(z, y)   $\Rightarrow$ relative(x, y) &           40 &        0.351014 \\
  mother(y, z) $\wedge$ relative(x, z)   $\Rightarrow$ relative(x, y) &           36 &        0.350490 \\
       child(z, x) $\wedge$ spouse(z, y)   $\Rightarrow$ father(x, y) &           48 &        0.350117 \\
 relative(z, x) $\wedge$ sibling(y, z)   $\Rightarrow$ relative(x, y) &           40 &        0.348765 \\
   child(y, z) $\wedge$ relative(x, z)   $\Rightarrow$ relative(x, y) &           37 &        0.343180 \\
  relative(x, z) $\wedge$ spouse(z, y)   $\Rightarrow$ relative(x, y) &           37 &        0.305043 \\
  relative(x, z) $\wedge$ spouse(y, z)   $\Rightarrow$ relative(x, y) &           37 &        0.305043 \\
  father(z, y) $\wedge$ relative(z, x)   $\Rightarrow$ relative(x, y) &           37 &        0.301061 \\
  father(y, z) $\wedge$ relative(z, x)   $\Rightarrow$ relative(x, y) &           37 &        0.294304 \\
  mother(z, y) $\wedge$ relative(x, z)   $\Rightarrow$ relative(x, y) &           33 &        0.283544 \\
   child(z, y) $\wedge$ relative(z, x)   $\Rightarrow$ relative(x, y) &           37 &        0.275454 \\
   child(y, z) $\wedge$ relative(z, x)   $\Rightarrow$ relative(x, y) &           37 &        0.273356 \\
  mother(y, z) $\wedge$ relative(z, x)   $\Rightarrow$ relative(x, y) &           33 &        0.241645 \\
  mother(z, y) $\wedge$ relative(z, x)   $\Rightarrow$ relative(x, y) &           30 &        0.224490 \\
  relative(z, x) $\wedge$ spouse(y, z)   $\Rightarrow$ relative(x, y) &           35 &        0.206044 \\
  relative(z, x) $\wedge$ spouse(z, y)   $\Rightarrow$ relative(x, y) &           32 &        0.205761 \\
    father(z, x) $\wedge$ father(y, z)   $\Rightarrow$ relative(x, y) &           30 &        0.196078 \\
  father(z, x) $\wedge$ relative(y, z)   $\Rightarrow$ relative(x, y) &           33 &        0.179144 \\
     child(z, y) $\wedge$ father(z, x)   $\Rightarrow$ relative(x, y) &           26 &        0.166667 \\
     child(x, z) $\wedge$ father(y, z)   $\Rightarrow$ relative(x, y) &           27 &        0.153584 \\
   child(x, z) $\wedge$ relative(y, z)   $\Rightarrow$ relative(x, y) &           32 &        0.148287 \\
  father(z, x) $\wedge$ relative(z, y)   $\Rightarrow$ relative(x, y) &           30 &        0.140871 \\
      child(z, y) $\wedge$ child(x, z)   $\Rightarrow$ relative(x, y) &           25 &        0.130759 \\
 relative(z, y) $\wedge$ sibling(x, z)   $\Rightarrow$ relative(x, y) &           39 &        0.127978 \\
 relative(z, y) $\wedge$ sibling(z, x)   $\Rightarrow$ relative(x, y) &           39 &        0.126881 \\
    father(z, x) $\wedge$ spouse(y, z)   $\Rightarrow$ relative(x, y) &           21 &        0.124101 \\
   child(x, z) $\wedge$ relative(z, y)   $\Rightarrow$ relative(x, y) &           31 &        0.123643 \\
    father(z, x) $\wedge$ spouse(z, y)   $\Rightarrow$ relative(x, y) &           17 &        0.122083 \\
 relative(y, z) $\wedge$ sibling(z, x)   $\Rightarrow$ relative(x, y) &           38 &        0.120448 \\
 relative(y, z) $\wedge$ sibling(x, z)   $\Rightarrow$ relative(x, y) &           38 &        0.120360 \\
    father(z, y) $\wedge$ father(x, z)   $\Rightarrow$ relative(x, y) &           16 &        0.108774 \\
  mother(x, z) $\wedge$ relative(y, z)   $\Rightarrow$ relative(x, y) &           21 &        0.104492 \\
     child(x, z) $\wedge$ spouse(y, z)   $\Rightarrow$ relative(x, y) &           13 &        0.102020 \\
     child(x, z) $\wedge$ spouse(z, y)   $\Rightarrow$ relative(x, y) &           13 &        0.102020 \\
  mother(z, x) $\wedge$ relative(y, z)   $\Rightarrow$ relative(x, y) &           18 &        0.101064 \\
   father(x, z) $\wedge$ sibling(z, y)   $\Rightarrow$ relative(x, y) &           10 &        0.100971 \\
relative(z, y) $\wedge$ relative(x, z)   $\Rightarrow$ relative(x, y) &           29 &        0.100440 \\
\bottomrule
\caption[Rules mined from the original family KG]{Rules mined from the original family KG, with their corresponding PCA confidences and how many times they were mined from the 48 extensions. Note that all rules with the $relative$ predicate in the consequent have lower PCA confidence and were mined less frequently.}%
\label{family_original_rules_table_PCA}
\end{longtable}

\end{document}